\pdfoutput=1

\RequirePackage{snapshot}
\documentclass[11pt]{article}

\usepackage{acl}
\usepackage{times}
\usepackage{latexsym}

\usepackage{amsfonts}

\usepackage{microtype}

\usepackage[subtle]{savetrees}
\usepackage{multirow}
\usepackage{hyperref}
\usepackage{booktabs}
\usepackage{tabularx}
\usepackage{graphicx}
\usepackage{subcaption}
\usepackage[htt]{hyphenat}
\usepackage{sidecap}
\usepackage[T1]{fontenc}
\usepackage{ragged2e}
\usepackage{siunitx}
\usepackage{adjustbox}
\usepackage{amsmath,amsfonts,bm}
\usepackage{mathtools}
\newcommand*\rot{\rotatebox{90}}
\newcommand{\STAB}[1]{\begin{tabular}{@{}c@{}}\rot{#1}\end{tabular}}

\usepackage{xurl}
\usepackage{xspace}

\usepackage{algorithm,algpseudocode}

\usepackage{times}
\usepackage{latexsym}
\usepackage{amsfonts,amsthm,amssymb}
\usepackage{amsmath}
\usepackage{euscript}
\usepackage{amstext}
\usepackage{graphicx}
\usepackage{color}
\usepackage{url}
\usepackage{verbatim}
\usepackage{xspace}
\usepackage{framed}
\usepackage{multirow}

\newtheorem{lemma}{Lemma}[section]

\newtheorem{definition}[lemma]{Definition}

{\hspace*{\fill}$\Box$\par}

	\newcommand{\initOneLiners}{%
		\setlength{\itemsep}{0pt}
		\setlength{\parsep }{pt}
		\setlength{\topsep }{0pt}
	}

	\newcommand{\cD}{\mathcal{D}}

\usepackage{fullpage}
\usepackage{amsthm,amsmath,amssymb}
\usepackage{mdframed}
\usepackage{graphicx}
\usepackage{enumitem}
\usepackage{xspace}
\usepackage{mdframed}
\usepackage{multirow}
\usepackage{hhline}

\makeatletter
\renewcommand{\paragraph}{%
	\@startsection{paragraph}{4}%
	{\z@}{1.25ex \@plus 1ex \@minus .2ex}{-1em}%
	{\normalfont\normalsize\bfseries}%
}
\makeatother

\renewcommand{\tilde}{\widetilde}

\algnewcommand{\To}{\textbf{To }}
\algnewcommand\Input{\item[\textbf{Input:}]}%
\algnewcommand\Output{\item[\textbf{Output:}]}%

\usepackage[disable]{todonotes}
\makeatletter
\newcommand*\iftodonotes{\if@todonotes@disabled\expandafter\@secondoftwo\else\expandafter\@firstoftwo\fi}
\makeatother
\newcommand{\noindentaftertodo}{\iftodonotes{\noindent}{}\ignorespaces}

\newcommand{\Fixme}[2][]{\noindentaftertodo}
\newcommand{\Notewho}[3][]{\noindentaftertodo}
\newcommand{\Jason}[2][]{\noindentaftertodo}
\newcommand{\Ysu}[2][]{\noindentaftertodo}
\newcommand{\Richard}[2][]{\noindentaftertodo}
\newcommand{\Fatemeh}[2][]{\noindentaftertodo}

\usepackage[T1]{fontenc}

\usepackage[utf8]{inputenc}

\usepackage{microtype}
\usepackage{listings}
\usepackage{titlesec}
\titlespacing*{\section}{0.5ex}{1ex}{1ex}
\titlespacing*{\subsection}{1ex}{1ex}{.5ex}
\titlespacing*{\subsubsection}{0pt}{.2ex}{.2ex}

\newcommand{\puser}{p_{\mathrm{priv}}}
\newcommand{\Duser}{D_{\mathrm{priv}}}
\newcommand{\Dpub}{D_{\mathrm{pub}}}
\newcommand{\Dsynth}{D_{\mathrm{s}}}
\newcommand{\plp}{p_{\phi_0}(y \mid x)}

\title{Privacy-Preserving Domain Adaptation of Semantic Parsers}

\author{Fatemehsadat Mireshghallah\textsuperscript{\rm 1,\rm 2}\thanks{\quad Work done as part of a Microsoft Semantic Machines internship. Corresponding author email: fatemeh@ucsd.edu} \quad  Yu Su\textsuperscript{\rm 2}\\
    \textbf{Tatsunori Hashimoto}\textsuperscript{\rm 2} \quad \textbf{Jason Eisner}\textsuperscript{\rm 2} \quad \textbf{Richard Shin}\textsuperscript{\rm 2}\\
    \textsuperscript{\rm 1} University of California, San Diego 
    \textsuperscript{\rm 2} Microsoft Semantic Machines\\
    {\small \texttt{fatemeh@ucsd.edu
    \{yusu2,v-hashimotot,jason.eisner,richard.shin\}@microsoft.com}}\\
    }

\begin{document}
\maketitle
\begin{abstract}
Task-oriented dialogue systems often assist users with personal or confidential matters. For this reason, the developers of such a system are generally prohibited from observing actual usage.  So how can they know where the system is failing and needs more training data or new functionality?
In this work, we study ways in which realistic user utterances can be generated synthetically, to help increase the linguistic and functional coverage of the system, without compromising the privacy of actual users. 
To this end, we propose a two-stage Differentially Private (DP) generation method which first generates latent semantic parses,
and then generates utterances based on the parses. 
Our proposed approach improves MAUVE by 2.5$\times$ and parse tree function type overlap by 1.3$\times$ relative to current approaches for private synthetic data generation, improving both on fluency and semantic coverage. We further validate our approach on a realistic domain adaptation task of adding new functionality from private user data to a semantic parser, and show overall gains of $8.5\%$ points in accuracy with the new feature.

\end{abstract}

\section{Introduction}

\begin{figure*}[t!]
    \centering
\includegraphics[width=0.95\linewidth]{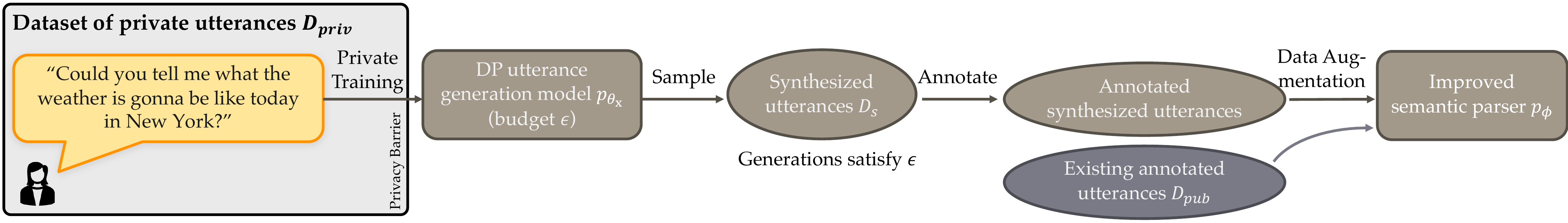}
     \caption{Overview of our problem setup and a baseline approach: We want to improve the performance of a semantic parser using differentially private synthesized utterances, based on private user data.}
     \label{fig:1stage}
    \vspace{-1.0ex}
\end{figure*}

In task-oriented dialogue systems, such as Siri and Alexa, a software agent parses a user’s intent into a program, executes it and then communicates the results back to the user~\cite{andreas2020task,parse1,cheng-etal-2020-conversational,gupta2018semantic,pomdp}.
As a result of their growing popularity, these systems face an increasing demand to improve their linguistic coverage (\emph{How do users talk?\@}) as well as functional coverage (\emph{What are users trying to do?}).
An input utterance to such a system could look like this: ``Could you tell me what the weather is gonna be like today in New York?'' and the agent answers it by predicting and executing a program, or \textit{semantic parse}.

In many cases, due to privacy controls, system developers can only use datasets that are limited and contrived, e.g., dialogues created by crowd workers pretending to be users. This is a significant domain shift from real private user data.
Unlabeled data from real
user interactions with dialogue systems has abundant signals that could be used to improve the linguistic and functional coverage of semantic parsers. For instance, practitioners could detect gaps in coverage by examining interactions where the system was unsure about what to do or responded that it lacked the requested functionality.

Note that training on real user interactions would be problematic for privacy even if automated and unsupervised.
Trained models can 
``memorize'' details of their training data~\cite{tirumala2022memorization,feldman,mireshghallah-etal-2022-empirical}, and this can be exploited through different types of attacks that either extract full training sequences  from models~\cite{carlini2021extracting,carlini2019secret} or infer the presence of a given sequence of interest in the training data~\cite{mireshghallah2022quantifying}.

To mitigate that problem, Differentially Private (DP) training algorithms, such as DP-SGD~\cite{abadi2016deep,DworkKMMN06}, can be used to provide worst-case guarantees on the information leakage of a trained model. This guarantee is controlled by the privacy budget $\epsilon$, where lower epsilon means higher privacy. But while DP-SGD could be used to adapt (fine-tune) a semantic parser on unannotated private data, there is a limit to what can be done in this way.  Even if some users are asking the system to hop up and down, fine-tuning is unlikely to make it grow legs.  Thus, our goal in this paper is to use DP-SGD to produce realistic data that can be inspected (so that the developers know to build legs) and expertly annotated (to rapidly teach the semantic parser that words like ``hop'' and ''jump'' should invoke the leg API).  

Recent concurrent work~\cite{yue2022synthetic,mattern2022differentially} has attempted to generate synthetic labeled data to train a text classifier.
They sample labels from a given distribution and sample corresponding text strings from a differentially private language model, conditioning each string on its label. 
This approach does not directly apply to our setting, as our distribution over labels is private and one of our goals is to learn it. We cannot even substitute a uniform distribution for the sake of generation, since no such distribution exists over the infinitely many labels of the semantic parsing task.

To meet the unique requirements of our setting,
a simple baseline is to ignore the classes and simply train a differentially private language model on all the private utterances, applying DP-SGD to the usual log-likelihood objective
(Figure~\ref{fig:1stage}).
We could then sample synthetic utterances for inspection and annotation.  However, we find that when we enforce privacy with budget $\epsilon=3$, this baseline's top-25 most common function types\footnote{Function types are the set of unique, discrete node labels in the semantic parse tree, corresponding to API methods.} only have $64\%$ overlap with the top-25 most common function types in the private utterances (see Section~\ref{sec:overall}  and Table~\ref{tab:break-coverage} for full experimental details and results).  In other words, the baseline model does not accurately capture the  distribution of the private training data, over a limited number of synthesized samples.

\begin{figure*}[t!]
    \centering
\includegraphics[width=0.95\linewidth]{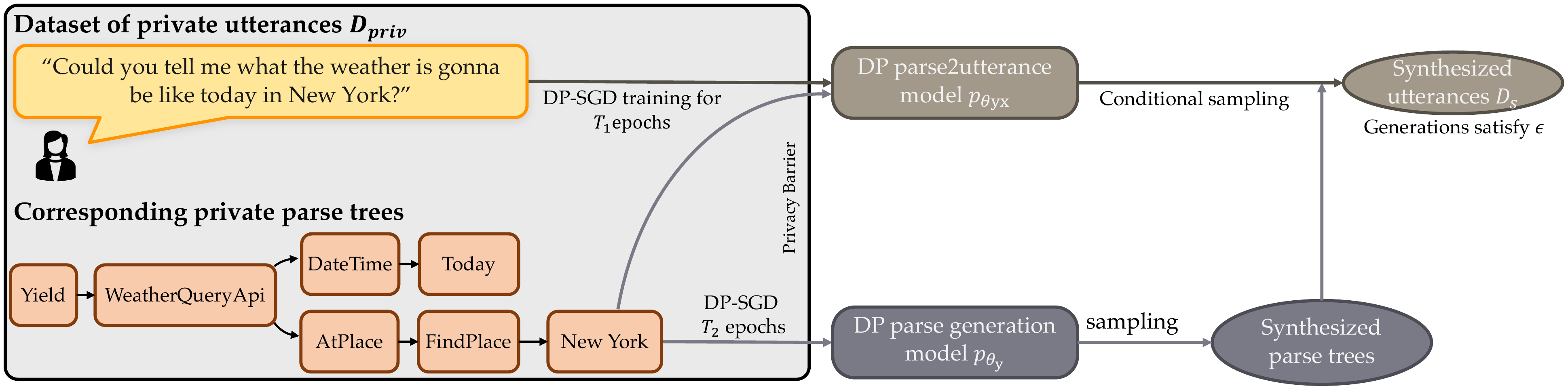}
     \caption{Overview of the proposed 2-stage method: On the left, we see an example of a private user utterance with its (also private) semantic parse tree.  On the right we see how our 2-stage framework trains DP parse tree generation and parse2utterance models, and then samples from them to produce synthesized utterances. }
     \label{fig:2stage}
    \vspace{-1ex}
\end{figure*}

To ensure sufficient coverage of how users invoke different function types, we propose a 2-stage method (Figure~\ref{fig:2stage}) that exploits the structure of the output space, by privately (using DP-SGD) modeling the parse trees (bottom of the figure) and the conditional distribution of the utterances given a parse tree (top of the figure), separately. These models can then be used to generate as many samples as desired, by first sampling parse trees from the parse generation model and then prompting the parse2utterance model with these parse trees.
Using the proposed method, we observe a $80\%$ coverage of the top-25 most common function types in private utterances, which is a significant improvement over the baseline ($64\%$).
To further evaluate the efficacy of our method in improving a downstream system, we annotate DP-generated utterances to add a missing functionality to a low-resource semantic parser, 
and show that the parser's accuracy on this missing functionality is $13.4\%$ higher if the DP-generated utterances come from our 2-stage method rather than the 1-stage baseline. 

\section{Background}\label{sec:main-background}
\begin{definition}[Differential Privacy (DP) \citep{DworkKMMN06}]
  A randomized algorithm $\mathcal{A}$ is  ($\epsilon$,$\delta$)-differentially private if for any two neighboring inputs $D$ and $D'$, which differ in exactly the data pertaining to a single record, and for any set $\mathcal{S}$ of possible outputs: 
$
\textstyle{\Pr[\mathcal{A}(D) \in \mathcal{S}] \leq e^{\epsilon}\,\Pr[\mathcal{A}(D') \in \mathcal{S}] +\delta}
$.
\end{definition}

To train a neural network with differential privacy, the most widely used algorithm is the  DP variant of stochastic gradient descent, DP-SGD~\citep{abadi2016deep}. It resembles ordinary SGD, but at each gradient update step, it first clips the per-example gradient to a maximum norm of $C$, then obfuscates it by adding Gaussian noise with mean $0$ and standard deviation $\sigma C$.
Intuitively, this limits the contribution that any single example makes to the final model parameters returned by the training algorithm $\mathcal{A}$.
The privacy expenditure of DP-SGD, $(\epsilon, \delta)$, is a function of $C$, $\sigma$, $|B|$ (batch size), $|\cD|$ (dataset size), and the total number of epochs $T$ (which controls the total number of gradient updates). It is determined based on the R\'enyi DP~\cite{mironov2017renyi} privacy accounting method. In practice, following prior work, we fine-tune our models using DP-Adam~\cite{abadi2016deep,li2021large,yu2021differentially}. We elaborate more on DP training in Appendix~\ref{app:dp-sgd}.

\noindent\textbf{Post-processing property.} This property of DP ensures that if an algorithm $\mathcal{A}$ satisfies
$(\epsilon, \delta)$-DP, then so does  $F \circ \mathcal{A}$ for any function $F$~\cite{DworkKMMN06}.  This means that we can  take as many samples as we want from the DP-trained models, without changing the privacy expenditure.

\begin{algorithm}[H]
  \begin{algorithmic}[1]
        \Input{Utterance generation model $p_{\theta_\text{x}(x)}$, private utterances $\Duser$,  privacy budget $(\epsilon, \delta)$}, batch size $|B|$, epochs $T$, clipping threshold $C$, privacy accountant $\mathcal{A}$
        \State Feed the parameters $\epsilon$, $\delta$, $C$, $T$, and $|B|$ to the accountant $\mathcal{A}$ to get noise multiplier $\sigma$
        \State Fine-tune $p_{\theta_x}(x)$ on $\Duser$ with DP-SGD and parameters $\sigma$ and $C$ for $T$ epochs with batch size $|B|$
        \State Populate $\Dsynth$ with samples from  $p_{\theta_\text{x}}$
        \State \Return $\Dsynth$
  \end{algorithmic}
    \caption{Differentially Private Training and Sampling: 1-stage Baseline}
    \label{alg:1stage}
\end{algorithm}
\section{Method}
\label{sec:method}

\noindent\textbf{Setting.}
We are given an unlabeled private dataset $\Duser = \{x_i\}$ drawn from the distribution of private user utterances, $\puser(x)$.  We are also given a labeled public dataset $\Dpub = \{(x_j,y_j)\}$, and a semantic parser $p_{\phi_0}(y \mid x)$ already trained on $\Dpub$.

\noindent\textbf{Goal.} Our goal is to find an $(\epsilon,\delta)$-DP model $\phi$, such that $p_{\phi}(y\mid x)$ has a lower loss than $p_{\phi_0}(y\mid x)$ on the task of semantically parsing utterances from $\puser$.
We will achieve this by proposing different methods of using $\Duser$ with DP to train a model $p_\theta$ of $p_\text{priv}$, and using it to synthesize an unlabeled dataset, $\Dsynth$, that can then be manually annotated and used to augment the training set $\Dpub$ for learning $\phi$.

\paragraph{}
In the rest of this section, we first introduce the baseline method for $p_\theta(x)$, 
then we propose our 2-stage method.
In both cases, the output of the process is a dataset of unlabeled synthetic utterances, $\Dsynth$, which is similar to $\Duser$ and will later be annotated and used to augment $\Dpub$ when training $\phi$.
Due to the post-processing property (see Section~\ref{sec:main-background}), any model trained on $\Dsynth$ still satisfies $(\epsilon,\delta)$-DP.

\begin{algorithm}[H]
  \begin{algorithmic}[1]
        \Input{
        Parse tree generation $p_{\theta_\text{y}}(y)$  and parse2utterance $p_{\theta_\text{yx}}(x\mid y)$ models,
        private utterances $\Duser$, privacy budget $(\epsilon, \delta)$}, batch size $|B|$, $T_1$ and $T_2$ as  epochs allocated to stages 1 and 2, clipping threshold $C$, privacy accountant $\mathcal{A}$, low-resource parser $\plp$ trained on $\Dpub$
         \State Feed the parameters $\epsilon$, $\delta$, $C$, $T=T_1+T_2$, and $|B|$ to the accountant $\mathcal{A}$ to get noise multiplier $\sigma$
        \State Feed $\Duser$ to $p_{\phi_0}$ to sample a parse tree for each utterance, and augment $\Duser$ with the trees
        \State 
        Fine-tune $p_{\theta_\text{y}}(y)$ on only \textit{parse trees} from $\Duser$ with DP-SGD and parameters $\sigma$ and $C$ for  $T_1$ epochs with batch size $|B|$
        \State Fine-tune $p_{\theta_\text{yx}}(x \mid y)$ on $\Duser$ with DP-SGD and parameters $\sigma$ and $C$ for  $T_2$ epochs
        \State Take samples from $p_{\theta_\text{y}}(y)$ with batch size $|B|$
        \State Prompt $p_{\theta_\text{yx}}(x \mid y)$ with the  sampled parse trees from the previous step, and populate  $\Dsynth$ with the output samples.
      \State  \Return $\Dsynth$
  \end{algorithmic}
    \caption{Differentially Private Training and Sampling: Proposed 2-stage Technique}
    \label{alg:2stage}
\end{algorithm}

\subsection{Baseline: Vanilla DP Language Model}\label{sec:baseline}

Figure~\ref{fig:1stage} shows the 1-stage baseline approach of fine-tuning a pre-trained generative auto-regressive language model on private user utterances using DP-SGD~\cite{abadi2016deep,li2021large,yu2021differentially}.
Algorithm~\ref{alg:1stage} details this process:
To create the synthesized dataset, we first fine-tune the initial utterance model $p_{\theta_x}$ on $\Duser$, using DP-SGD with noise multiplier $\sigma$, batch size $|B|$ and clipping threshold $C$, for $T$ epochs. The noise multiplier is given to us by the privacy accountant (see Section~\ref{sec:background}).
Then, we take samples from this fine-tuned model. Due to the post-processing property of DP, any sample from this model satisfies the guarantees  and \emph{the synthesized dataset size does not affect the privacy guarantees.}

\subsection{Proposed Method: 2-stage Generation}\label{sec:2stage}

As we discuss in the introduction and the results sections, the baseline fails to accurately capture the distribution of different word-types (vocabulary) and parse tree function-types in the private utterances.
To increase the linguistic and functional coverage of the synthesized data, we propose a 2-stage method to exploit the inherent structure in the private user-utterance parse trees.
We assume that the parse trees for the private utterances are also private, since they are almost unique to each utterance, and cannot be released/used without privacy measures. 
We also assume that the parse trees are noisy, e.g., generated by low-resource parsers rather than expert annotators, and we use such trees in our end-to-end experiments.
Figure~\ref{fig:2stage} shows an overview of our proposed 2-stage generation model, with the bottom part showing (in blue) one stage of the process, training a parse tree generation model, with DP-SGD, to model the parse trees as intermediate (latent) variables. 
A model like $p(x)= \sum_y p(y) p(x|y)$ is a latent variable model (for example, a variational autoencoder) of $x$, where we can sample $x$, by first drawing $y$, then drawing $x$ given $y$, then discarding $y$. That is what we have done, with $y$ being a parse-tree and $x$ being an utterance.
The other stage is training a parse2utterance model  that would take the intermediate variables (parse trees, $y$) as input prompts and produce the utterance corresponding to them.
It is noteworthy that \textit{the training of these two stages is completely independent} and can be parallelized.

Algorithm~\ref{alg:2stage} shows the details of the training and sampling process. 
An important design choice is how to split $T$, the number of overall training epochs, between the two stages as $T=T_1+T_2$. 
This effectively splits the privacy budget: the stage that gets more epochs consumes more of the privacy budget. We discuss this further and run ablation studies in Section~\ref{sec:hyperparams}.

We choose to use the same privacy parameters ($\sigma$ and $C$) determined from $T$ for training of both stages, as this enables us to use the sophisticated privacy accountant of R\'enyi DP (RDP)~\cite{mironov2017renyi} and achieve a tighter bound on the privacy parameter $\epsilon$, compared to directly splitting $\epsilon$ and using different privacy parameters for each stage. The privacy accountant keeps track of how much privacy budget $\epsilon$ (information that an adversary can recover about a training sample) has been spent so far during the training process. We elaborate more on this choice in Appendix~\ref{app:privacy}.

\section{Experimental Setup}
In this section we briefly describe our experimental setup. For full  experimental details see App.~\ref{app:exp-details}.

\subsection{Datasets}
\label{sec:dataset}

We use two large-scale conversational semantic parsing datasets, SMCalFlow v2.0~\cite{andreas2020task} and TreeDST~\cite{cheng-etal-2020-conversational}. We pre-process them to break the conversations with the agents into single turns, each consisting of an (utterance, parse tree) pair, and we only use the human turns. This yields a training/test dataset with size of 133,584/14,571 and 121,652/22,897 for SMCalFlow and TreeDST respectively. 

\subsection{Metrics}\label{sec:metrics}

We compare the synthesized datasets to human utterances on a distribution level, as there is no one-to-one mapping between them. We report two sets of metrics here: (1)~Language and (2)~Parse metrics.  Language metrics (MAUVE and word overlap) are measured from the generated utterances themselves, whereas parse metrics (chi-square distance and function overlap) are measured from parse trees corresponding to the synthesized utterances, produced using a high-resource parser.\footnote{A parser trained on a large amount of data, close to SotA accuracy.}  All the results are reported over $14,751$ and $22,897$ synthesized utterances, compared to the same number of human utterances, for SMCalFlow and TreeDST datasets respectively.

\noindent\textbf{MAUVE}~\cite{mauve2021} is a comparison measure for open-ended text generation, which directly compares the learned distribution from a text generation model to the distribution of human-written text using divergence frontiers.\footnote{ ~\url{https://github.com/krishnap25/mauve}}

\noindent\textbf{Word-type Overlap} (W.\@ Overlap) measures the word-type (vocabulary) coverage of  generated text, as the ratio of overlapping types between the generated text and human utterances, against the human utterances, when the text is tokenized using spaces.

\noindent\textbf{Function Type Overlap} (F.\@ Overlap)  measures the function type coverage of the parse trees of the synthesized text, as the ratio of overlapping parse tree API function types between the generated text and human utterances, against the human utterances.

\noindent\textbf{Chi-square Distance} (Dist.\@) measures the 
$\chi^2$
distance between the distribution of API function types from the parse trees of the synthesized utterances against the  parse trees of human utterances.

\subsection{Model Architectures and Decoding}\label{sec:arch}

\noindent\textbf{Semantic Parser.}
We train a Transformer-based semantic parser from~\citet{zhou2022online} on only the human turns from the dialogues (without any context). Details are provided in Appendix~\ref{app:hparams}.

\noindent\textbf{1-stage and 2-stage models.} We fine-tune a pre-trained GPT-2 (small) from Hugging Face~\cite{wolf2019huggingface} for all three of the utterance generation, parse tree generation, and parse2utterance models. We also provide results for GPT2-Large in the Appendix~\ref{app:large}.

\noindent\textbf{Decoding.} We use Hugging Face's multinomial beam search (\texttt{beam\_sample}) with beam width of 1 for decoding from the parse-generation model and beam width of 5 for decoding from parse2utterance and utterance generation models, as found in our hyperparameter search. 

\subsection{Baselines}\label{sec:all_baselines}

\noindent\textbf{1-stage.} This baseline denotes the method explained in Section~\ref{sec:baseline}, where we fine-tune a pre-trained language model (GPT2-Small) with DP-SGD~\cite{abadi2016deep}, on the utterances. This baseline is essentially equivalent to the methods proposed in~\citet{li2021large,yu2021differentially}.

\noindent\textbf{1-stage + domain (1.5-stage).} For further evaluation of our method, we devise a more sophisticated baseline, inspired by~\citet{yue2022synthetic}, where we augment the 1-stage model with a constrained set of prompts that reflect the domain of the modeled utterance. As such, we fine-tune  GPT2-Small with DP-SGD to create a domain2utterance model on the TreeDST dataset, which uses the domain label (10 domains: flight, hotel, etc.) instead of the parse tree.  We sample domains from the true domain distribution. This is also similar to the setup in~\citet{mattern2022differentially}, though they only target classification tasks.

\section{Experimental Results}

In this section we first compare the baseline and our proposed method.  We then study possible reasons for the observed superiority of the proposed 2-stage technique, and analyze hyperparameter sensitivity for the generation process. Finally, we compare the performance of the baseline and the proposed method on improving the performance of a downstream semantic parser. 
We provide additional experiments and ablations (such as hyperparameter sensitivity and detailed result break-downs) in Appendix~\ref{app:additional} alongside sample synthesized utterances and parse-trees in Tables~\ref{tab:generations} and~\ref{tab:trees_gens}, as a reference.

\subsection{Comparison with Baselines}\label{sec:overall}

Table~\ref{tab:overview} shows a comparison of the 1-stage and 1-stage + domain (1.5 stage) baselines (Section~\ref{sec:all_baselines}) with our proposed 2-stage method, for three different privacy budgets of $\epsilon=\{3,8,16\}$, for the two datasets SMCalFlow and TreeDST.
We present results for the 1-stage + domain  baseline only on the TreeDST dataset, since we have the domain annotations only for this dataset.
The ground-truth row reports the metrics for the utterances in the test set, which is  why the language metrics are both at the maximum value $1.0$. However, the parse metrics are not perfect since the high-resource parser used for evaluations does not achieve $100\%$ accuracy on the test set.

We can see that the proposed 2-stage method outperforms both the 1-stage baselines, at all levels of privacy budget, even when the privacy budget is $\infty$ (i.e., the No DP row in the table), for both datasets.
The 1-stage + domain baseline has a performance that is on average better than the 1-stage baseline (hypothethically due to the guidance that the domain prompts provide) but inferior to the proposed 2-stage method.
We can also observe that on average, as the privacy budget increases (lower privacy), the performance of all methods increases, which makes sense as the added noise is decreasing.

Both single and 2-stage methods perform better overall on the TreeDST dataset in terms of the parse metrics. This could be due to the smaller set of function types for TreeDST (303 for TreeDST vs.\ 524 for SMCalFlow), making it easier for both methods to capture these types.
One counter-intuitive observation is that for some metrics, for the 1-stage baseline, the performance with $\epsilon=16$ is actually higher than $\epsilon=\infty$. We hypothesize that this could be due to the regularization effect that small amounts of noise has on the training~\cite{smith2020generalization}, therefore the DP model with a high budget can generalize better (overfits less). We also observe that for the SMCalFlow dataset, MAUVE doesn't improve much as we increase the privacy budget. We relate this to the complexity of the hyperparameter search/optimization in DP-mechanisms, and that we were not optimizing for improving MAUVE. A more extensive hyperparameter search could yield better results on $\epsilon=16$.

We provide more fine-grained comparisons on the parse tree distribution matching with ground truth (such as the top-10, 25, 50 and 100 most common function coverage in Appendix~\ref{app:coverage}).
We explore the reasons behind the superior performance of the proposed method in the next section.

\begin{table*}[]
    \centering
    \vspace{-2ex}
    \begin{adjustbox}{width=\linewidth, center}
    
 \newcolumntype{L}{>{\RaggedLeft\arraybackslash}p{0.06\linewidth}} 
  \newcolumntype{O}{>{\RaggedLeft\arraybackslash}m{0.07\linewidth}} 
  \newcolumntype{D}{>{\arraybackslash}m{0.15\linewidth}} 
  \newcolumntype{R}{>{\arraybackslash}m{0.29\linewidth}} 
\begin{tabular}{@{}clcccccccccccccccc@{}}
	\toprule
	&  {\multirow{3}{*}{}} &\multicolumn{5}{c}{SMCalFlow} &{} &\multicolumn{5}{c}{TreeDST}\\
 	\cmidrule{3-7} \cmidrule{9-13}
 & & \multicolumn{2}{c}{{Language Metrics}}&{}& \multicolumn{2}{c}{{Parse Metrics
 }} &{} & \multicolumn{2}{c}{{Language Metrics}}&{}& \multicolumn{2}{c}{{Parse Metrics
 }}   \\
	\cmidrule{3-4} \cmidrule{6-7}  	\cmidrule{9-10} \cmidrule{12-13}
	&  Method &  {W.\@ Overlap $\uparrow$}  &{Mauve $\uparrow$}&{} & {Dist.\@ $\downarrow$} & {F.\@ Overlap $\uparrow$} & {}& {W.\@ Overlap $\uparrow$}  &{Mauve $\uparrow$}&{} & {Dist.\@ $\downarrow$} & {F.\@ Overlap $\uparrow$}  \\
    \midrule
    \multirow{1}{*}{\STAB{}} 
    &  G.\@ Truth & 1.0	& 1.0 &	&0.003&	0.820&& 1.0	& 1.0 &	&0.001&	0.990	\\
    \midrule[0.1pt] 
   \multirow{2}{*}{\STAB{\textcolor{black}{No DP}}} 
    &  1-stage       &0.087$\pm$0.005&0.334$\pm$0.056&&0.258$\pm$0.034&0.487$\pm$0.004& &0.125$\pm$0.021&0.274$\pm$0.043&&0.101$\pm$0.019&0.702$\pm$0.033	\\
    & 1.5-stage &-&-&&-&-&&0.230$\pm$0.005&0.399$\pm$0.046&&0.079$\pm$0.010&0.842$\pm$0.025 \\
    & 2-stage        &\textbf{0.236}$\pm$0.012&\textbf{0.632}$\pm$0.005&&\textbf{0.085}$\pm$0.009&\textbf{0.797}$\pm$0.006& &\textbf{0.456}$\pm$0.012&\textbf{0.521}$\pm$0.019&&\textbf{0.040}$\pm$0.006&\textbf{0.966}$\pm$0.004  \\
           \midrule[0.1pt]
    \multirow{2}{*}{\STAB{$\epsilon=16$}}
    &  1-stage    &0.092$\pm$0.012&0.258$\pm$0.073&&0.167$\pm$0.018&0.499$\pm$0.025 &&0.142$\pm$0.015&0.213$\pm$0.041&&0.108$\pm$0.016&0.755$\pm$0.031	    \\
    & 1.5-stage &-&-&&-&-&&0.174$\pm$0.008&0.176$\pm$0.031&&0.102$\pm$0.019&0.774$\pm$0.051 \\
    & 2-stage     &\textbf{0.213}$\pm$0.007&\textbf{0.524}$\pm$0.027&&\textbf{0.057}$\pm$0.003&\textbf{0.708}$\pm$0.013 &&\textbf{0.290}$\pm$0.013&\textbf{0.384}$\pm$0.021&&\textbf{0.035}$\pm$0.005&\textbf{0.954}$\pm$0.009     \\
           \midrule[0.1pt]
    \multirow{2}{*}{\STAB{$\epsilon=8$}}
    &  1-stage          &0.093$\pm$0.014&0.198$\pm$0.053&&0.183$\pm$0.011&0.487$\pm$0.052&&0.140$\pm$0.016&0.199$\pm$0.032&&0.110$\pm$0.023&0.773$\pm$0.027 	\\
    & 1.5-stage &-&-&&-&-&&0.172$\pm$0.009&0.178$\pm$0.017&&0.102$\pm$0.024&0.793$\pm$0.041 \\
    & 2-stage           &\textbf{0.210}$\pm$0.007&\textbf{0.533}$\pm$0.032&&\textbf{0.055}$\pm$0.004&\textbf{0.707}$\pm$0.010&&\textbf{0.281}$\pm$0.018&\textbf{0.354}$\pm$0.047&&\textbf{0.036}$\pm$0.006&\textbf{0.945}$\pm$0.010   \\
           \midrule[0.1pt]
    \multirow{2}{*}{\STAB{$\epsilon=3$}}
    &  1-stage          &0.086$\pm$0.016&0.138$\pm$0.035&&0.185$\pm$0.001&0.485$\pm$0.059&&0.138$\pm$0.023&0.176$\pm$0.030&&0.111$\pm$0.030&0.795$\pm$0.058  	\\
    & 1.5-stage &-&-&&-&-&&0.166$\pm$0.006&0.147$\pm$0.016&&0.105$\pm$0.024&0.801$\pm$0.016 \\
    & 2-stage           &\textbf{0.205}$\pm$0.004&\textbf{0.530}$\pm$0.031&&\textbf{0.054}$\pm$0.003&\textbf{0.693}$\pm$0.010&&\textbf{0.256}$\pm$0.008&\textbf{0.294}$\pm$0.040&&\textbf{0.037}$\pm$0.005&\textbf{0.938}$\pm$0.007    \\
	\bottomrule
\end{tabular}

    \end{adjustbox}
   \caption{Comparison of the proposed 2-stage method with the 1-stage and the 1-stage + domain (1.5 stage) baseline, with different levels of privacy budget ($\epsilon$), where lower budget means higher privacy. The numbers are presented as mean $\pm\sigma$ (unbiased sample standard deviation), over three runs with three different seeds.}
       \label{tab:overview}
    \vspace{-2ex}
\end{table*}

\subsection{Ablation Studies}
We hypothesize that the superiority of the 2-stage method, which models the parse trees as intermediate variables, is because it (1) improves the language modeling within each utterance and (2) helps the model learn the different semantic modes in the data. 

\paragraph{Disrupting the Correlation between Parse-trees and Utterances.}
We first ~\textit{disrupt the correlation between parse trees and utterances by shuffling them} (i.e., each utterance is now paired with a random parse tree). We discuss the full details and results of this experiment in Appendix~\ref{app:shuffle}, but in short we observe that in this setup, the parse-related metric (function overlap) for 2-stage synthetic data falls from $63.9\%$ to $23.2\%$, which is below that of the 1-stage baseline ($47.5\%$), supporting the hypothesis that the structure in the parse trees and the correlation to utterances are important.
Based on this, in the rest of this section, we test our multi-modality hypothesis by limiting the data to fewer modes.

\noindent\textbf{Changing the Modes in the Data.}
Our conjecture is that part of why the 2-stage model benefits from explicitly learning a distribution over semantic parses is that this helps it capture the different semantic modes in the data---that is, the various types of functionality invoked by the utterances.

To test this hypothesis,
we create a subsample of the original dataset, \textit{consisting only of (utterance, parse tree) pairs where the parse tree contains the Weather function}.
This ``single-mode'' dataset focuses on weather-related queries.
We compare the pattern of results to that in the original experiment, where the dataset had greater diversity of function types.
Note that, due to the high compositionality of the parse trees (e.g., a parse tree that contains the Weather function may also contain many other functions for, e.g., datetimes and locations),  the restricted dataset still contains $158$ function types, compared to the $524$ in the  original data.

Table~\ref{tab:modes} shows the results for this experiment, and compares the performance of the 1-stage and 2-stage methods (the numbers don't  match those of Table~\ref{tab:overview} for the same $\epsilon$ value, as for this experiment we use a smaller batch size of $1024$ for the sake of run-time). As we can see, the improvement achieved by the 2-stage method shrinks on the restricted dataset, which supports our conjecture.
It is noteworthy that although the improvement shrinks, it remains relatively high for the metrics that consider parses, showing that the 2-stage method retains an advantage in capturing the remaining functional diversity in the restricted dataset. 

\begin{table}[]
    \centering
    \vspace{-2ex}
    \begin{adjustbox}{width=0.99\linewidth, center}
     \newcolumntype{L}{>{\RaggedLeft\arraybackslash}p{0.06\linewidth}} 
  \newcolumntype{O}{>{\RaggedLeft\arraybackslash}m{0.07\linewidth}} 
  \newcolumntype{D}{>{\arraybackslash}m{0.15\linewidth}} 
  \newcolumntype{R}{>{\arraybackslash}m{0.29\linewidth}} 
\begin{tabular}{@{}cclcc@{}}

	\toprule
	&&  Method &  {MAUVE $\uparrow$}&{Dis. $\downarrow$}  \\
    \midrule
    \multirow{1}{*}{\STAB{}} 
   \multirow{4}{*}{{\STAB{No DP}}} &\multirow{2}{*}{\textcolor{black}{Few-modes}} 
    &  1-stage     & 0.234$\pm$0.023&0.241$\pm$0.024	\\
    && 2-stage     & 0.214$\pm$0.062&0.105$\pm$0.032    \\
        \cmidrule{2-5}
    &\multirow{2}{*}{{Full-modes}}
    &  1-stage     &0.334$\pm$0.056&0.258$\pm$0.034	\\
    && 2-stage     &0.632$\pm$0.005&0.085$\pm$0.009  \\

\midrule
   \multirow{4}{*}{{\STAB{$\epsilon$=8}}} &\multirow{2}{*}{{Few-modes}} 
    &  1-stage      &0.285$\pm$0.063&0.194$\pm$0.014   \\
    && 2-stage      &0.224$\pm$0.037&0.169$\pm$0.039   \\
        \cmidrule{2-5}
    &\multirow{2}{*}{{\textcolor{black}{Full-modes}}}
    &  1-stage      &0.203$\pm$0.103&0.163$\pm$0.019   \\
    && 2-stage      &0.428$\pm$0.046&0.073$\pm$0.010   \\

	\bottomrule
\end{tabular}

    \end{adjustbox}
    \caption{The effect of using  few-modal data for training vs.\ the full dataset, on the performance of the 1-stage baseline and the proposed 2-stage method. The goal here is to see if the superiority of the 2-stage method is due to it better capturing different modes in the data. The numbers are presented as mean $\pm$ standard deviation, over three runs with different random seeds.}
        \label{tab:modes}
    \vspace{-2ex}
\end{table}

\subsection{Hyperparameter Sensitivity Analysis}\label{sec:hyperparams}

We run extensive analysis to study the effect of different hyperparameters (batchsize, learning rate, clipping threshold and the total number of the epochs for the 2-stage method) on the quality of the synthesized text. For the sake of space we present all these results in Appendix~\ref{app:hparams}.
In short, we find that as the batch size increases, the quality of the generated text also improves, which has been observed by prior work in DP generation of text ~\cite{li2021large}. For splitting of the privacy budget between the training of the parse-generation and  parse2utterance models, we find that most of the epochs should be allocated to the latter: 
increasing $T_2$ at the expense of $T_1$ steadily improves the quality of the generated text (under both text-based and parse-based metrics), until a tipping point is reached.  We find $T_1=2$ and $T_2=8$ epochs to be the best setup.

\subsection{Downstream Parser Improvement}\label{sec:end-to-end}\label{sec:downstream-semantic-parser-improvement}

\begin{table*}[]
    \centering
    \vspace{-2ex}
    \begin{adjustbox}{width=\linewidth, center}

 \newcolumntype{L}{>{\RaggedLeft\arraybackslash}p{0.06\linewidth}} 
  \newcolumntype{O}{>{\RaggedLeft\arraybackslash}m{0.07\linewidth}} 
  \newcolumntype{D}{>{\arraybackslash}m{0.15\linewidth}} 
  \newcolumntype{R}{>{\arraybackslash}m{0.29\linewidth}} 
\begin{tabular}{@{}clccccccccc@{}}

	\toprule
	& {\multirow{2}{*}{}} &  \multicolumn{3}{c}{{Missing (Weather/EventOnDate) function  breakdown
 }}&{}& \multicolumn{3}{c}{{Average over all function types}}   \\
	\cmidrule{3-5} \cmidrule{7-9}
	&  Method &  Anonymized Graph Match&	API Precision	&API Recall&{} & Anonymized Graph Match&	API Precision	& API Recall \\
    \midrule
    \multirow{1}{*}{\STAB{}} 
&Full Dataset	&76.6 / 74.0&77.5 / 78.5 &77.5 / 78.6&&73.9$\pm$0.1&78.1$\pm$0.3&78.2$\pm$0.4\\	
    \midrule[0.1pt] 
   \multirow{3}{*}{\STAB{\footnotesize{Weather}}} 
&Non-augmented	&0.0$\pm$0.0&1.9$\pm$0.7&2.1$\pm$0.9    && 60.9$\pm$1.1&65.7$\pm$1.2&65.6$\pm$1.3\\					    
&  1-stage      &37.7$\pm$2.6&42.1$\pm$3.1&42.1$\pm$3.1 && 65.9$\pm$0.2&71.0$\pm$0.5&71.0$\pm$0.5	 	    \\
& 2-stage       &43.7$\pm$0.7&50.3$\pm$0.9&50.3$\pm$0.9 && 66.4$\pm$0.6&71.9$\pm$0.4&71.9$\pm$0.4	      \\
    \midrule[0.1pt] 
   \multirow{3}{*}{\STAB{\footnotesize{EoD}}} 
&Non-augmented	 &0.0$\pm$0.0&4.4$\pm$0.1&4.3$\pm$0.0    && 57.6$\pm$0.3&61.4$\pm$0.9&61.4$\pm$0.9 \\					    
&  1-stage       &48.0$\pm$0.7&52.2$\pm$1.1&51.9$\pm$1.1 && 64.0$\pm$0.5&67.9$\pm$0.3&67.9$\pm$0.3    		 	    \\
& 2-stage        &61.4$\pm$0.7&65.6$\pm$0.7&65.3$\pm$0.5 && 66.1$\pm$0.5&71.0$\pm$0.5&70.9$\pm$0.5    		      \\

	\bottomrule
\end{tabular}

    \end{adjustbox}
    \caption{End-to-end experiment results (low-resource semantic parser augmentation). The Weather and EoD rows determine the composition of $\Dpub$ (Section~\ref{sec:end-to-end}). The ``Missing (Weather/EventOnDate)'' columns report the metrics over only the functionality that was missing from the public data, but present in $\Duser$ (since Full Dataset isn't missing a function we report both functions, separated by `/'). The ``Average over all'' columns report metric over all function types.
}
         \label{tab:end-to-end-main}
    \vspace{-2ex}
\end{table*}

In this section, we demonstrate a major application of our privacy-preserving data synthesis, through an end-to-end experiment: improving the performance of a low-resource semantic parser and adding new functionality to it based on private user data. 
Building on the notation described in Section~\ref{sec:method},
we assume access to a small ``eyes-on'' (non-private) dataset $\Dpub$ of (utterance, parse tree) pairs, and a semantic parser $p_{\phi_0}(y \mid x)$ trained on $\Dpub$.
We also assume we have ``eyes-off'' access to a much larger unlabeled private utterance dataset $\Duser$,
which we can only access through a DP mechanism.
Our goal is to synthesize dataset $\Dsynth$ such that a parser trained on $\Dpub \cup \Dsynth$ performs better on dataset $\Duser$ than the original parser $p_{\phi_0}(y \mid x)$. 

We devise this experiment such that $\Dpub$ is missing function types that $\Duser$ has, and we aim to capture this missing functionality through the DP data synthesis and augmentation. Specifically, we will construct $\Dpub$ by removing one class  of function from SMCalFlow: either \textit{Weather} or \textit{event on date} (EoD). We chose these two to be distinct, as weather-related queries comprise only $3.4\%$ of the samples in the dataset, whereas EoD appears in $10.7\%$. Comparing these two helps us study the effect of a function's commonness on the ability to add it through this procedure.

\noindent\textbf{Dataset Partitioning.}
We uniformly subsample $\frac{1}{10}$  of the pre-processed SMCalFlow training set (from Section~\ref{sec:dataset}) to form $\Dpub$. Each entry contains a human-generated utterance and the corresponding expert-annotated parse tree. We drop all the samples with weather or EoD from $\Dpub$. We use the remaining $\frac{9}{10}$th of the dataset to form $\Duser$. We test on a uniformly sampled subset of the test data (with the same $\frac{1}{10}$ ratio).

\noindent\textbf{Parser Metrics.} We measure the performance of the initial low-resource semantic parser and the augmented ones using the following three metrics, adopted from \citet{zhou2022online}: (1)~\textit{Exact Anonymized Graph Match}, which reports the percentage of test samples for which the anonymized generated parse tree is an exact match to the ground truth expert annotations from SMCalFlow; (2\&3)~\textit{API Match Precision and Recall}, which measures the precision and recall of the generated parse tree nodes (API functions) from the parser, treating the tree as a bag of nodes, against those of the ground truth.

\noindent\textbf{Experimental Procedure.}

Starting with a low-resource parser $p_{\phi_0}(y \mid x)$ trained on $\Dpub$, and we aim to improve it using $\Duser$. We first obtain a predicted parse tree for each utterance in $\Duser$ by running it through $\plp$. We then train the 1-stage baseline and our 2-stage method on $\Duser$ with DP-SGD (we set $\epsilon=3$), and then take samples from them to form $\Dsynth$, as outlined in Algorithms~\ref{alg:1stage} and~\ref{alg:2stage}. For the purpose of this experiment, we set the size of $\Dsynth$ to $90,000$ samples~\footnote{As we want to makes $\Dpub \cup \Dsynth$ have about the same size as the original SMCalFlow training set). The future work can explore different ways to vary exactly how we sample $\Dsynth$ (such as focusing on the fenced utterances, or changing the size of $\Dsynth$).}. At this point, when applied in practice, we would have experts annotate the utterances in $\Dsynth$ with their parse trees. For this experiment, lacking human annotators, we use a high-resource parser (Appendix~\ref{app:hparams}) to predict a parse tree as a reasonably good approximation to the ground truth. Finally, we augment $\Dpub$ with $\Dsynth$, re-train
 the semantic parser, and compare its performance to $\plp$.

\noindent\textbf{Results.} Table~\ref{tab:end-to-end-main} shows the results for the different compositions of $\Dpub$, described earlier in this section. We provide the language and parse metrics (from the previous sections) for these generations in Appendix~\ref{app:e2ebreak}, alongside a more fine-grained breakdown of the results in Appendix Table~\ref{tab:end-to-end-app}. 
As expected,  the exact graph match performance of the non-augmented parser when $\Dpub$ is missing  Weather or EoD is $0.0$ on utterances containing those functions (the precision and recall are not exactly zero since there are queries that contain Weather alongside other function types and those other types are correctly identified). 
After augmentation, we see that both methods for synthesizing utterances $\Dsynth$ lead to parser improvements, with the 2-stage method providing more overall improvement. 

It is on EoD that we observe the most improvement (and the most pronounced gap between the two methods), especially from the 2-stage method.  Presumably this is due to the higher prevalence of EoD functions in $\Duser$, so there are more training samples ($10.7\%$ vs.\@ the $3.4\%$ of weather) for it than the private training could use. We conjecture that this is also the reason behind the bigger performance gap between the single and two stage models for this function, as  there  are more samples wrongly annotated by the low-resource parser, and the 2-stage method is picking up on this through its use of the parse trees.

Another observation is that the gap between all the augmentations and using the full dataset is still quite significant. We believe this is due to the fact that we used the low-resource parser to provide annotations for the private data, which means inaccurate annotations are being fed to the 2-stage method for training. Therefore $\Duser$ is still far from the fully expert annotated data used to train the parser in the first row of the table. Using a better parser, or iteratively augmenting and re-annotating, might help close this gap by providing more accurate parse trees to the 2-stage method.

\section{Related Work}
\label{sec:related-work}

We offer a brief summary of related work here.  For a detailed discussion see Appendix~\ref{app:related-work}.

\noindent\textbf{PII scrubbing.}
Techniques such as automated removal of personally identifiable information (PII)~\cite{lison-etal-2021-anonymisation,pii1} and training with redacted data~\cite{zhao2022provably} are often used to protect user privacy, especially in medical settings~\cite{kayaalp2014identification}. However, alone, they do not provide stringent guarantees or bounds on information leakage~\cite{Brown2022WhatDI}, as there are many forms of private information not captured in PII removal. Rather, DP is used as a gold standard for limiting leakage.

\noindent\textbf{Differentially Private Data Synthesis.} Recent concurrent work~\cite{yue2022synthetic,mattern2022differentially} has attempted different variations of taking samples from a DP-trained model~\cite{abadi2016deep,li2021large,yu2021differentially,kerrigan-etal-2020-differentially,shi2021selective,anil2021large,tian2021seqpate} to synthesize labeled data for classification tasks, by conditioning the generation on the label of each sample, and assuming that the prior distribution over labels is known and public. They then use this data to augment and improve classification models.
In our case, however, we are not dealing with finite labels or classification, we want to improve the performance of a semantic parser, which is a structured (hierarchical) task, where there are infinite possible parses that we cannot enumerate, and the labels are also  private (since parse trees are almost unique we consider them private). Therefore, existing methods cannot be applied to this problem, as mentioned in the introduction.

Another difference in this concurrent line of work is their reliance on gold (ground truth) labels. In our case, as we show in the final experiment, we can instead train our DP models on imperfect parse trees that are generated by low-resource parsers.

\noindent\textbf{Semantic Parsing.}
The closest work to ours is \citet{yang-etal-2022-addressing}, which deals with the problem of safely learning from private user utterances, starting with a  low-resource semantic parser. Unlike our setup, however, they do not rely on DP and are only concerned with the removal of PII. While they also consider a distribution over parse trees and employ a parse-tree-to-utterance process, they implement the latter using only the original supervised data, which forecloses the possibility of adapting to distributional shifts as in our experiments. 
Another line of work advocates for attenuating privacy risks by enabling semantic parsers to autonomously learn from interacting with users~\cite{yao-etal-2020-imitation,karamcheti2020learning,yao-etal-2019-model}.
Through privacy-preserving data synthesis, our method supports diagnostics of private user traffic and better control over the associated learning process.

\noindent\textbf{Data Augmentation}
Data augmentation for performance improvement is also relevant to our work.~\citet{1,2,3,4,5}  propose data generation methods designed for the ``intents and slots'' model where each utterance is considered to have one of a fixed set of intents, and each intent has a fixed set of slots, each of which needs to be filled with a value. We, in contrast, use the SMCalFlow and TreeDST datasets which use compositional semantic representations of arbitrary complexity.
~\citet{2} and~\citet{4} propose to use automated paraphrasing to create semantically equivalent variants of an existing utterance. Each utterance is paraphrased independently of the others.  Unfortunately, paraphrases of a private utterance cannot be made public as they leak information about the private utterance; obtaining differentially private paraphrases would require new research.  Training the paraphrasing model by DP-SGD will not help (and is generally unnecessary as such a model can be trained on public data).

\section{Conclusions}

In this paper, we studied the problem of using private user data to improve semantic parser performance in task-oriented dialogue systems, without violating user privacy. We proposed a two-stage method for differentially private utterance synthesis that exploits the inherent structure in the parse trees to better fit the private distribution. We showed that this method outperforms a baseline DP generative language model on a variety of datasets and metrics.
We also demonstrated the effectiveness of our method in an end-to-end application scenario where we improved the performance of a low-resource parser by adding new functionality that was motivated by private user data. We showed that our method provided   overall gains of $8.5\%$ points in accuracy with the new feature.

\section*{Limitations}

DP training of large models is compute-intensive, requiring per-example gradients and large batch sizes~\cite{li2021large,subramani2021enabling}.  This renders the training of such models difficult and not easily accessible to everyone. 

DP-SGD takes records to be single training examples, which in this paper's experiments correspond to single user utterances.  That setup prevents the trained model from revealing much information about any given single utterance, but it may still allow information to leak that is repeated across multiple utterances \cite{Brown2022WhatDI}.  

For both the baseline method and our two-stage method, we trained our model to approximately match the true distribution of private user utterances, $\puser(x)$, to the extent that this was possible under a differential privacy guarantee.  Of course, there are many ways to measure the quality of an approximation, and different approximations are appropriate for different tasks where it might be important to preserve different properties of $\puser(x)$. The one-stage baseline approach implicitly aims to achieve a low cross-entropy, by applying DP-SGD to the log-likelihood function.  In contrast, our two-stage approach aims to encourage an approximation that also roughly preserves the marginal distribution over semantic function types.  We did not investigate more direct ways of encouraging such an approximation, for example, one-stage DP-SGD with a modified objective function that explicitly evaluates the marginal distribution in addition to the log-likelihood.  

Finally, 
we trained an approximate model of $\puser$ from which we can draw utterances to inspect and annotate.  But we must acknowledge that $\puser$ is not the ideal distribution to approximate.  Even if we were able to actually use private utterances to improve the system, we would not necessarily want to draw them directly from $\puser$.  Rather, we would want to select them by active learning---selecting the private user utterances that would be \emph{most useful} to inspect or to include in the annotated training data.  Thus, when training our model by DP-SGD (using either the one-stage or two-stage procedure), we could upweight or upselect the private utterances that appear useful in this way---resulting in a differentially private model that generates \emph{useful} synthetic utterances.  Specifically, traditional active learning by uncertainty sampling \cite{settles-2012} would select utterances where the semantic parser was uncertain what to do.  We would also want to select utterances where the system suspected for other reasons that it did not do the right thing---because it classified the user's request as a functionality that the system did not yet support, or because the user objected in some way to the system's response.  We have left experiments on this setup to future work.

\section*{Ethics Statement}
 The over-arching goal of our work is to improve semantic parsers and dialogue systems while protecting the privacy rights of users who contribute their data to this goal.
While we train our models by applying DP mechanisms with worst-case guarantees, deploying these models in real-world setups and using these synthesized data-sets requires further verification that users’ privacy is preserved, by setting the right definition of ``record'' (i.e., training example) and the right $(\epsilon,\delta)$ budget based on privacy policy guidelines.  Further studies are needed on what the reported privacy budgets actually mean in practice for users, how users perceive these privacy mechanisms, and how they can provide informed consent to have their data used to improve the systems~\cite{Brown2022WhatDI}.
\bibliography{acl_latex}

\begin{thebibliography}{54}
\expandafter\ifx\csname natexlab\endcsname\relax\def\natexlab#1{#1}\fi

\bibitem[{Abadi et~al.(2016)Abadi, Chu, Goodfellow, McMahan, Mironov, Talwar,
  and Zhang}]{abadi2016deep}
Martin Abadi, Andy Chu, Ian Goodfellow, H~Brendan McMahan, Ilya Mironov, Kunal
  Talwar, and Li~Zhang. 2016.
\newblock Deep learning with differential privacy.
\newblock In \emph{Proceedings of the 2016 ACM SIGSAC conference on computer
  and communications security}, pages 308--318.

\bibitem[{Andreas et~al.(2020)Andreas, Bufe, Burkett, Chen, Clausman, Crawford,
  Crim, DeLoach, Dorner, Eisner et~al.}]{andreas2020task}
Jacob Andreas, John Bufe, David Burkett, Charles Chen, Josh Clausman, Jean
  Crawford, Kate Crim, Jordan DeLoach, Leah Dorner, Jason Eisner, et~al. 2020.
\newblock Task-oriented dialogue as dataflow synthesis.
\newblock \emph{Transactions of the Association for Computational Linguistics},
  8:556--571.

\bibitem[{Anil et~al.(2021)Anil, Ghazi, Gupta, Kumar, and
  Manurangsi}]{anil2021large}
Rohan Anil, Badih Ghazi, Vineet Gupta, Ravi Kumar, and Pasin Manurangsi. 2021.
\newblock Large-scale differentially private bert.
\newblock \emph{arXiv preprint arXiv:2108.01624}.

\bibitem[{Aura et~al.(2006)Aura, Kuhn, and Roe}]{pii1}
Tuomas Aura, Thomas~A. Kuhn, and Michael Roe. 2006.
\newblock \href {https://doi.org/10.1145/1179601.1179608} {Scanning electronic
  documents for personally identifiable information}.
\newblock In \emph{Proceedings of the 5th ACM Workshop on Privacy in Electronic
  Society}, WPES '06, page 41–50, New York, NY, USA. Association for
  Computing Machinery.

\bibitem[{Brown et~al.(2022)Brown, Lee, Mireshghallah, Shokri, and
  Tram{\`e}r}]{Brown2022WhatDI}
Hannah Brown, Katherine Lee, FatemehSadat Mireshghallah, R.~Shokri, and Florian
  Tram{\`e}r. 2022.
\newblock What does it mean for a language model to preserve privacy?
\newblock In \emph{Proceedings of the 2022 ACM Conference on Fairness,
  Accountability, and Transparency (FAccT)}.

\bibitem[{Burnyshev et~al.(2021)Burnyshev, Malykh, Bout, Artemova, and
  Piontkovskaya}]{burnyshev2021single}
Pavel Burnyshev, Valentin Malykh, Andrey Bout, Ekaterina Artemova, and Irina
  Piontkovskaya. 2021.
\newblock A single example can improve zero-shot data generation.
\newblock \emph{arXiv preprint arXiv:2108.06991}.

\bibitem[{Cao et~al.(2020)Cao, Zhu, Yang, Liu, Ma, Zhao, Chen, and
  Yu}]{cao2020unsupervised}
Ruisheng Cao, Su~Zhu, Chenyu Yang, Chen Liu, Rao Ma, Yanbin Zhao, Lu~Chen, and
  Kai Yu. 2020.
\newblock Unsupervised dual paraphrasing for two-stage semantic parsing.
\newblock \emph{arXiv preprint arXiv:2005.13485}.

\bibitem[{Carlini et~al.(2019)Carlini, Liu, Erlingsson, Kos, and
  Song}]{carlini2019secret}
Nicholas Carlini, Chang Liu, {\'U}lfar Erlingsson, Jernej Kos, and Dawn Song.
  2019.
\newblock The secret sharer: Evaluating and testing unintended memorization in
  neural networks.
\newblock In \emph{28th USENIX Security Symposium (USENIX Security 19)}, pages
  267--284.

\bibitem[{Carlini et~al.(2021)Carlini, Tramer, Wallace, Jagielski,
  Herbert-Voss, Lee, Roberts, Brown, Song, Erlingsson
  et~al.}]{carlini2021extracting}
Nicholas Carlini, Florian Tramer, Eric Wallace, Matthew Jagielski, Ariel
  Herbert-Voss, Katherine Lee, Adam Roberts, Tom Brown, Dawn Song, Ulfar
  Erlingsson, et~al. 2021.
\newblock Extracting training data from large language models.
\newblock In \emph{30th USENIX Security Symposium (USENIX Security 21)}, pages
  2633--2650.

\bibitem[{Cheng et~al.(2020)Cheng, Agrawal, Mart{\'\i}nez~Alonso, Bhargava,
  Driesen, Flego, Kaplan, Kartsaklis, Li, Piraviperumal, Williams, Yu,
  {\'O}~S{\'e}aghdha, and Johannsen}]{cheng-etal-2020-conversational}
Jianpeng Cheng, Devang Agrawal, H{\'e}ctor Mart{\'\i}nez~Alonso, Shruti
  Bhargava, Joris Driesen, Federico Flego, Dain Kaplan, Dimitri Kartsaklis, Lin
  Li, Dhivya Piraviperumal, Jason~D. Williams, Hong Yu, Diarmuid
  {\'O}~S{\'e}aghdha, and Anders Johannsen. 2020.
\newblock \href {https://doi.org/10.18653/v1/2020.emnlp-main.651}
  {Conversational semantic parsing for dialog state tracking}.
\newblock In \emph{Proceedings of the 2020 Conference on Empirical Methods in
  Natural Language Processing (EMNLP)}, pages 8107--8117, Online. Association
  for Computational Linguistics.

\bibitem[{Cho et~al.(2019)Cho, Xie, and Campbell}]{2}
Eunah Cho, He~Xie, and William~M. Campbell. 2019.
\newblock \href {https://doi.org/10.18653/v1/W19-2306} {Paraphrase generation
  for semi-supervised learning in {NLU}}.
\newblock In \emph{Proceedings of the Workshop on Methods for Optimizing and
  Evaluating Neural Language Generation}, pages 45--54, Minneapolis, Minnesota.
  Association for Computational Linguistics.

\bibitem[{Dwork et~al.(2006)Dwork, Kenthapadi, McSherry, Mironov, and
  Naor}]{DworkKMMN06}
Cynthia Dwork, Krishnaram Kenthapadi, Frank McSherry, Ilya Mironov, and Moni
  Naor. 2006.
\newblock Our data, ourselves: Privacy via distributed noise generation.
\newblock In \emph{Proceedings of the 24th Annual International Conference on
  the Theory and Applications of Cryptographic Techniques}, EUROCRYPT '06,
  pages 486--503, Berlin, Heidelberg. Springer.

\bibitem[{Feldman(2020)}]{feldman}
Vitaly Feldman. 2020.
\newblock \href {https://doi.org/10.1145/3357713.3384290} {Does learning
  require memorization? a short tale about a long tail}.
\newblock In \emph{Proceedings of the 52nd Annual ACM SIGACT Symposium on
  Theory of Computing}, STOC 2020, page 954–959, New York, NY, USA.
  Association for Computing Machinery.

\bibitem[{Ginart et~al.(2022)Ginart, van~der Maaten, Zou, and
  Guo}]{ginart2022submix}
Antonio Ginart, Laurens van~der Maaten, James Zou, and Chuan Guo. 2022.
\newblock Submix: Practical private prediction for large-scale language models.
\newblock \emph{arXiv preprint arXiv:2201.00971}.

\bibitem[{Gupta et~al.(2018)Gupta, Shah, Mohit, Kumar, and
  Lewis}]{gupta2018semantic}
Sonal Gupta, Rushin Shah, Mrinal Mohit, Anuj Kumar, and Mike Lewis. 2018.
\newblock Semantic parsing for task oriented dialog using hierarchical
  representations.
\newblock \emph{arXiv preprint arXiv:1810.07942}.

\bibitem[{Jolly et~al.(2020)Jolly, Falke, Tirkaz, and Sorokin}]{3}
Shailza Jolly, Tobias Falke, Caglar Tirkaz, and Daniil Sorokin. 2020.
\newblock \href {https://doi.org/10.18653/v1/2020.coling-industry.2}
  {Data-efficient paraphrase generation to bootstrap intent classification and
  slot labeling for new features in task-oriented dialog systems}.
\newblock In \emph{Proceedings of the 28th International Conference on
  Computational Linguistics: Industry Track}, pages 10--20, Online.
  International Committee on Computational Linguistics.

\bibitem[{Kairouz et~al.(2015)Kairouz, Oh, and
  Viswanath}]{kairouz2015composition}
Peter Kairouz, Sewoong Oh, and Pramod Viswanath. 2015.
\newblock The composition theorem for differential privacy.
\newblock In \emph{International conference on machine learning}, pages
  1376--1385. PMLR.

\bibitem[{Karamcheti et~al.(2020)Karamcheti, Sadigh, and
  Liang}]{karamcheti2020learning}
Siddharth Karamcheti, Dorsa Sadigh, and Percy Liang. 2020.
\newblock Learning adaptive language interfaces through decomposition.
\newblock \emph{EMNLP 2020}, page~23.

\bibitem[{Kayaalp et~al.(2014)Kayaalp, Browne, Dodd, Sagan, and
  McDonald}]{kayaalp2014identification}
Mehmet Kayaalp, Allen~C Browne, Zeyno~A Dodd, Pamela Sagan, and Clement~J
  McDonald. 2014.
\newblock De-identification of address, date, and alphanumeric identifiers in
  narrative clinical reports.
\newblock In \emph{AMIA Annual Symposium Proceedings}, volume 2014, page 767.
  American Medical Informatics Association.

\bibitem[{Kerrigan et~al.(2020)Kerrigan, Slack, and
  Tuyls}]{kerrigan-etal-2020-differentially}
Gavin Kerrigan, Dylan Slack, and Jens Tuyls. 2020.
\newblock \href {https://doi.org/10.18653/v1/2020.privatenlp-1.5}
  {Differentially private language models benefit from public pre-training}.
\newblock In \emph{Proceedings of the Second Workshop on Privacy in NLP}, pages
  39--45, Online. Association for Computational Linguistics.

\bibitem[{Kim et~al.(2021{\natexlab{a}})Kim, Gopi, Kulkarni, and
  Yekhanin}]{kim2021differentially}
Kunho Kim, Sivakanth Gopi, Janardhan Kulkarni, and Sergey Yekhanin.
  2021{\natexlab{a}}.
\newblock Differentially private n-gram extraction.
\newblock \emph{Advances in Neural Information Processing Systems},
  34:5102--5111.

\bibitem[{Kim et~al.(2021{\natexlab{b}})Kim, Chang, and Lee}]{kim2021neuralwoz}
Sungdong Kim, Minsuk Chang, and Sang-Woo Lee. 2021{\natexlab{b}}.
\newblock Neuralwoz: Learning to collect task-oriented dialogue via model-based
  simulation.
\newblock \emph{arXiv preprint arXiv:2105.14454}.

\bibitem[{Li et~al.(2021)Li, Tramer, Liang, and Hashimoto}]{li2021large}
Xuechen Li, Florian Tramer, Percy Liang, and Tatsunori Hashimoto. 2021.
\newblock Large language models can be strong differentially private learners.
\newblock \emph{arXiv preprint arXiv:2110.05679}.

\bibitem[{Li et~al.(2022)Li, Guo, Liu, Lou, and Xie}]{parse1}
Zhenwen Li, Jiaqi Guo, Qian Liu, Jian-Guang Lou, and Tao Xie. 2022.
\newblock Exploring the secrets behind the learning difficulty of meaning
  representations for semantic parsing.
\newblock In \emph{Proceedings of the 2022 Conference on Empirical Methods in
  Natural Language Processing (EMNLP)}.

\bibitem[{Lison et~al.(2021)Lison, Pil{\'a}n, Sanchez, Batet, and
  {\O}vrelid}]{lison-etal-2021-anonymisation}
Pierre Lison, Ildik{\'o} Pil{\'a}n, David Sanchez, Montserrat Batet, and Lilja
  {\O}vrelid. 2021.
\newblock \href {https://doi.org/10.18653/v1/2021.acl-long.323} {Anonymisation
  models for text data: State of the art, challenges and future directions}.
\newblock In \emph{Proceedings of the 59th Annual Meeting of the Association
  for Computational Linguistics and the 11th International Joint Conference on
  Natural Language Processing (Volume 1: Long Papers)}, pages 4188--4203,
  Online. Association for Computational Linguistics.

\bibitem[{Malandrakis et~al.(2019)Malandrakis, Shen, Goyal, Gao, Sethi, and
  Metallinou}]{1}
Nikolaos Malandrakis, Minmin Shen, Anuj Goyal, Shuyang Gao, Abhishek Sethi, and
  Angeliki Metallinou. 2019.
\newblock \href {https://doi.org/10.18653/v1/D19-5609} {Controlled text
  generation for data augmentation in intelligent artificial agents}.
\newblock In \emph{Proceedings of the 3rd Workshop on Neural Generation and
  Translation}, pages 90--98, Hong Kong. Association for Computational
  Linguistics.

\bibitem[{Mattern et~al.(2022)Mattern, Jin, Weggenmann, Schoelkopf, and
  Sachan}]{mattern2022differentially}
Justus Mattern, Zhijing Jin, Benjamin Weggenmann, Bernhard Schoelkopf, and
  Mrinmaya Sachan. 2022.
\newblock Differentially private language models for secure data sharing.
\newblock \emph{arXiv preprint arXiv:2210.13918}.

\bibitem[{Mireshghallah et~al.(2022{\natexlab{a}})Mireshghallah, Goyal, Uniyal,
  Berg-Kirkpatrick, and Shokri}]{mireshghallah2022quantifying}
Fatemehsadat Mireshghallah, Kartik Goyal, Archit Uniyal, Taylor
  Berg-Kirkpatrick, and Reza Shokri. 2022{\natexlab{a}}.
\newblock Quantifying privacy risks of masked language models using membership
  inference attacks.
\newblock In \emph{Proceedings of the 2022 Conference on Empirical Methods in
  Natural Language Processing (EMNLP)}.

\bibitem[{Mireshghallah et~al.(2022{\natexlab{b}})Mireshghallah, Uniyal, Wang,
  Evans, and Berg-Kirkpatrick}]{mireshghallah-etal-2022-empirical}
Fatemehsadat Mireshghallah, Archit Uniyal, Tianhao Wang, David Evans, and
  Taylor Berg-Kirkpatrick. 2022{\natexlab{b}}.
\newblock \href {https://aclanthology.org/2022.emnlp-main.119} {An empirical
  analysis of memorization in fine-tuned autoregressive language models}.
\newblock In \emph{Proceedings of the 2022 Conference on Empirical Methods in
  Natural Language Processing}, pages 1816--1826, Abu Dhabi, United Arab
  Emirates. Association for Computational Linguistics.

\bibitem[{Mironov(2017)}]{mironov2017renyi}
Ilya Mironov. 2017.
\newblock R{\'e}nyi differential privacy.
\newblock In \emph{2017 IEEE 30th computer security foundations symposium
  (CSF)}, pages 263--275. IEEE.

\bibitem[{Okur et~al.(2022)Okur, Sahay, and Nachman}]{4}
Eda Okur, Saurav Sahay, and Lama Nachman. 2022.
\newblock \href {https://aclanthology.org/2022.lrec-1.437} {Data augmentation
  with paraphrase generation and entity extraction for multimodal dialogue
  system}.
\newblock In \emph{Proceedings of the Thirteenth Language Resources and
  Evaluation Conference}, pages 4114--4125, Marseille, France. European
  Language Resources Association.

\bibitem[{Pillutla et~al.(2021)Pillutla, Swayamdipta, Zellers, Thickstun,
  Welleck, Choi, and Harchaoui}]{mauve2021}
Krishna Pillutla, Swabha Swayamdipta, Rowan Zellers, John Thickstun, Sean
  Welleck, Yejin Choi, and Zaid Harchaoui. 2021.
\newblock \href
  {https://proceedings.neurips.cc/paper/2021/file/260c2432a0eecc28ce03c10dadc078a4-Paper.pdf}
  {Mauve: Measuring the gap between neural text and human text using divergence
  frontiers}.
\newblock In \emph{Advances in Neural Information Processing Systems},
  volume~34, pages 4816--4828. Curran Associates, Inc.

\bibitem[{Settles(2012)}]{settles-2012}
Burr Settles. 2012.
\newblock \href {https://doi.org/10.1007/978-3-031-01560-1} {\emph{Active
  Learning}}.
\newblock Synthesis Lectures on Artificial Intelligence and Machine Learning.
  Springer.

\bibitem[{Shi et~al.(2021)Shi, Cui, Li, Jia, and Yu}]{shi2021selective}
Weiyan Shi, Aiqi Cui, Evan Li, Ruoxi Jia, and Zhou Yu. 2021.
\newblock Selective differential privacy for language modeling.
\newblock \emph{arXiv preprint arXiv:2108.12944}.

\bibitem[{Smith et~al.(2020)Smith, Elsen, and De}]{smith2020generalization}
Samuel Smith, Erich Elsen, and Soham De. 2020.
\newblock On the generalization benefit of noise in stochastic gradient
  descent.
\newblock In \emph{International Conference on Machine Learning}, pages
  9058--9067. PMLR.

\bibitem[{Su and Yan(2017)}]{su-yan-2017-cross}
Yu~Su and Xifeng Yan. 2017.
\newblock \href {https://doi.org/10.18653/v1/D17-1127} {Cross-domain semantic
  parsing via paraphrasing}.
\newblock In \emph{Proceedings of the 2017 Conference on Empirical Methods in
  Natural Language Processing}, pages 1235--1246, Copenhagen, Denmark.
  Association for Computational Linguistics.

\bibitem[{Subramani et~al.(2021)Subramani, Vadivelu, and
  Kamath}]{subramani2021enabling}
Pranav Subramani, Nicholas Vadivelu, and Gautam Kamath. 2021.
\newblock Enabling fast differentially private sgd via just-in-time compilation
  and vectorization.
\newblock \emph{Advances in Neural Information Processing Systems},
  34:26409--26421.

\bibitem[{Tian et~al.(2021)Tian, Zhao, Huang, Wang, Zhang, and
  He}]{tian2021seqpate}
Zhiliang Tian, Yingxiu Zhao, Ziyue Huang, Yu-Xiang Wang, Nevin Zhang, and
  He~He. 2021.
\newblock Seqpate: Differentially private text generation via knowledge
  distillation.

\bibitem[{Tirumala et~al.(2022)Tirumala, Markosyan, Zettlemoyer, and
  Aghajanyan}]{tirumala2022memorization}
Kushal Tirumala, Aram~H Markosyan, Luke Zettlemoyer, and Armen Aghajanyan.
  2022.
\newblock Memorization without overfitting: Analyzing the training dynamics of
  large language models.
\newblock \emph{arXiv preprint arXiv:2205.10770}.

\bibitem[{Tseng et~al.(2021)Tseng, Dai, Kreyssig, and
  Byrne}]{tseng2021transferable}
Bo-Hsiang Tseng, Yinpei Dai, Florian Kreyssig, and Bill Byrne. 2021.
\newblock Transferable dialogue systems and user simulators.
\newblock \emph{arXiv preprint arXiv:2107.11904}.

\bibitem[{Wang et~al.(2015)Wang, Berant, and Liang}]{wang-etal-2015-building}
Yushi Wang, Jonathan Berant, and Percy Liang. 2015.
\newblock \href {https://doi.org/10.3115/v1/P15-1129} {Building a semantic
  parser overnight}.
\newblock In \emph{Proceedings of the 53rd Annual Meeting of the Association
  for Computational Linguistics and the 7th International Joint Conference on
  Natural Language Processing (Volume 1: Long Papers)}, pages 1332--1342,
  Beijing, China. Association for Computational Linguistics.

\bibitem[{Wolf et~al.(2019)Wolf, Debut, Sanh, Chaumond, Delangue, Moi, Cistac,
  Rault, Louf, Funtowicz et~al.}]{wolf2019huggingface}
Thomas Wolf, Lysandre Debut, Victor Sanh, Julien Chaumond, Clement Delangue,
  Anthony Moi, Pierric Cistac, Tim Rault, R{\'e}mi Louf, Morgan Funtowicz,
  et~al. 2019.
\newblock Huggingface's transformers: State-of-the-art natural language
  processing.
\newblock \emph{arXiv preprint arXiv:1910.03771}.

\bibitem[{Yang et~al.(2022)Yang, Deng, Chen, Shin, Roy, and
  Van~Durme}]{yang-etal-2022-addressing}
Kevin Yang, Olivia Deng, Charles Chen, Richard Shin, Subhro Roy, and Benjamin
  Van~Durme. 2022.
\newblock \href {https://doi.org/10.18653/v1/2022.findings-acl.291} {Addressing
  resource and privacy constraints in semantic parsing through data
  augmentation}.
\newblock In \emph{Findings of the Association for Computational Linguistics:
  ACL 2022}, pages 3685--3695, Dublin, Ireland. Association for Computational
  Linguistics.

\bibitem[{Yao et~al.(2019)Yao, Su, Sun, and Yih}]{yao-etal-2019-model}
Ziyu Yao, Yu~Su, Huan Sun, and Wen-tau Yih. 2019.
\newblock \href {https://doi.org/10.18653/v1/D19-1547} {Model-based interactive
  semantic parsing: A unified framework and a text-to-{SQL} case study}.
\newblock In \emph{Proceedings of the 2019 Conference on Empirical Methods in
  Natural Language Processing and the 9th International Joint Conference on
  Natural Language Processing (EMNLP-IJCNLP)}, pages 5447--5458, Hong Kong,
  China. Association for Computational Linguistics.

\bibitem[{Yao et~al.(2020)Yao, Tang, Yih, Sun, and
  Su}]{yao-etal-2020-imitation}
Ziyu Yao, Yiqi Tang, Wen-tau Yih, Huan Sun, and Yu~Su. 2020.
\newblock \href {https://doi.org/10.18653/v1/2020.emnlp-main.559} {An imitation
  game for learning semantic parsers from user interaction}.
\newblock In \emph{Proceedings of the 2020 Conference on Empirical Methods in
  Natural Language Processing (EMNLP)}, pages 6883--6902, Online. Association
  for Computational Linguistics.

\bibitem[{Yin et~al.(2022)Yin, Wieting, Sil, and
  Neubig}]{yin-etal-2022-ingredients}
Pengcheng Yin, John Wieting, Avirup Sil, and Graham Neubig. 2022.
\newblock \href {https://doi.org/10.18653/v1/2022.acl-long.103} {On the
  ingredients of an effective zero-shot semantic parser}.
\newblock In \emph{Proceedings of the 60th Annual Meeting of the Association
  for Computational Linguistics (Volume 1: Long Papers)}, pages 1455--1474,
  Dublin, Ireland. Association for Computational Linguistics.

\bibitem[{Young et~al.(2013)Young, Gašić, Thomson, and Williams}]{pomdp}
Steve Young, Milica Gašić, Blaise Thomson, and Jason~D. Williams. 2013.
\newblock \href {https://doi.org/10.1109/JPROC.2012.2225812} {Pomdp-based
  statistical spoken dialog systems: A review}.
\newblock \emph{Proceedings of the IEEE}, 101(5):1160--1179.

\bibitem[{Yu et~al.(2021)Yu, Naik, Backurs, Gopi, Inan, Kamath, Kulkarni, Lee,
  Manoel, Wutschitz et~al.}]{yu2021differentially}
Da~Yu, Saurabh Naik, Arturs Backurs, Sivakanth Gopi, Huseyin~A Inan, Gautam
  Kamath, Janardhan Kulkarni, Yin~Tat Lee, Andre Manoel, Lukas Wutschitz,
  et~al. 2021.
\newblock Differentially private fine-tuning of language models.
\newblock \emph{arXiv preprint arXiv:2110.06500}.

\bibitem[{Yue et~al.(2022)Yue, Inan, Li, Kumar, McAnallen, Sun, Levitan, and
  Sim}]{yue2022synthetic}
Xiang Yue, Huseyin~A Inan, Xuechen Li, Girish Kumar, Julia McAnallen, Huan Sun,
  David Levitan, and Robert Sim. 2022.
\newblock Synthetic text generation with differential privacy: A simple and
  practical recipe.
\newblock \emph{arXiv preprint arXiv:2210.14348}.

\bibitem[{Zhang et~al.(2022)Zhang, Jiang, Chen, Chen, and Zheng}]{5}
Xin Zhang, Miao Jiang, Honghui Chen, Chonghao Chen, and Jianming Zheng. 2022.
\newblock \href {https://doi.org/10.3390/math10183358} {Cloze-style data
  augmentation for few-shot intent recognition}.
\newblock \emph{Mathematics}, 10(18).

\bibitem[{Zhao et~al.(2022)Zhao, Li, and Wang}]{zhao2022provably}
Xuandong Zhao, Lei Li, and Yu-Xiang Wang. 2022.
\newblock Provably confidential language modelling.
\newblock \emph{arXiv preprint arXiv:2205.01863}.

\bibitem[{Zhao et~al.(2019)Zhao, Zhu, and Yu}]{zhao2019data}
Zijian Zhao, Su~Zhu, and Kai Yu. 2019.
\newblock Data augmentation with atomic templates for spoken language
  understanding.
\newblock \emph{arXiv preprint arXiv:1908.10770}.

\bibitem[{Zhong et~al.(2020)Zhong, Lewis, Wang, and
  Zettlemoyer}]{zhong2020grounded}
Victor Zhong, Mike Lewis, Sida~I Wang, and Luke Zettlemoyer. 2020.
\newblock Grounded adaptation for zero-shot executable semantic parsing.
\newblock \emph{arXiv preprint arXiv:2009.07396}.

\bibitem[{Zhou et~al.(2022)Zhou, Eisner, Newman, Platanios, and
  Thomson}]{zhou2022online}
Jiawei Zhou, Jason Eisner, Michael Newman, Emmanouil~Antonios Platanios, and
  Sam Thomson. 2022.
\newblock Online semantic parsing for latency reduction in task-oriented
  dialogue.
\newblock In \emph{Proceedings of the 60th Annual Meeting of the Association
  for Computational Linguistics (Volume 1: Long Papers)}, pages 1554--1576.

\end{thebibliography}
\bibliographystyle{acl_natbib}

\appendix

\clearpage
\label{sec:appendix}

\begin{figure}[t!]
    \centering
     \includegraphics[width=0.99\linewidth]{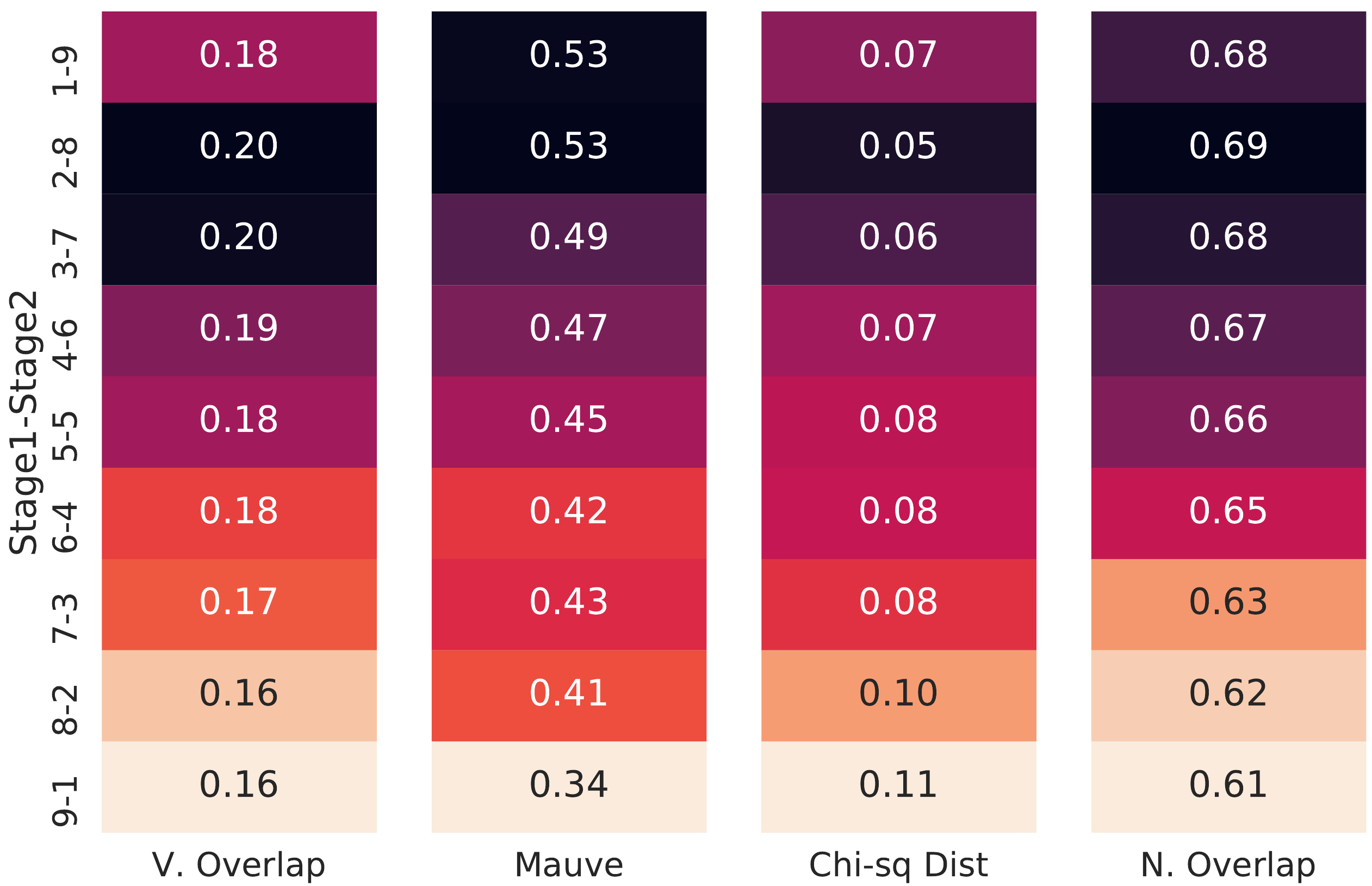}
     \caption{Study of privacy budget split between the two stages of the proposed method. We do the privacy accounting assuming an overall $10$ epochs of training for both stages together, and the budget is split by splitting these $10$ epochs between the two stages; the $y$ axis shows this. The first/second number is epochs spent on training the parse generation/parse2utterance model.}
     \label{fig:budget}
    \vspace{-1ex}
\end{figure}

\begin{figure*}[]
    \centering
    \begin{subfigure}{0.23\textwidth}
     \includegraphics[width=\linewidth]{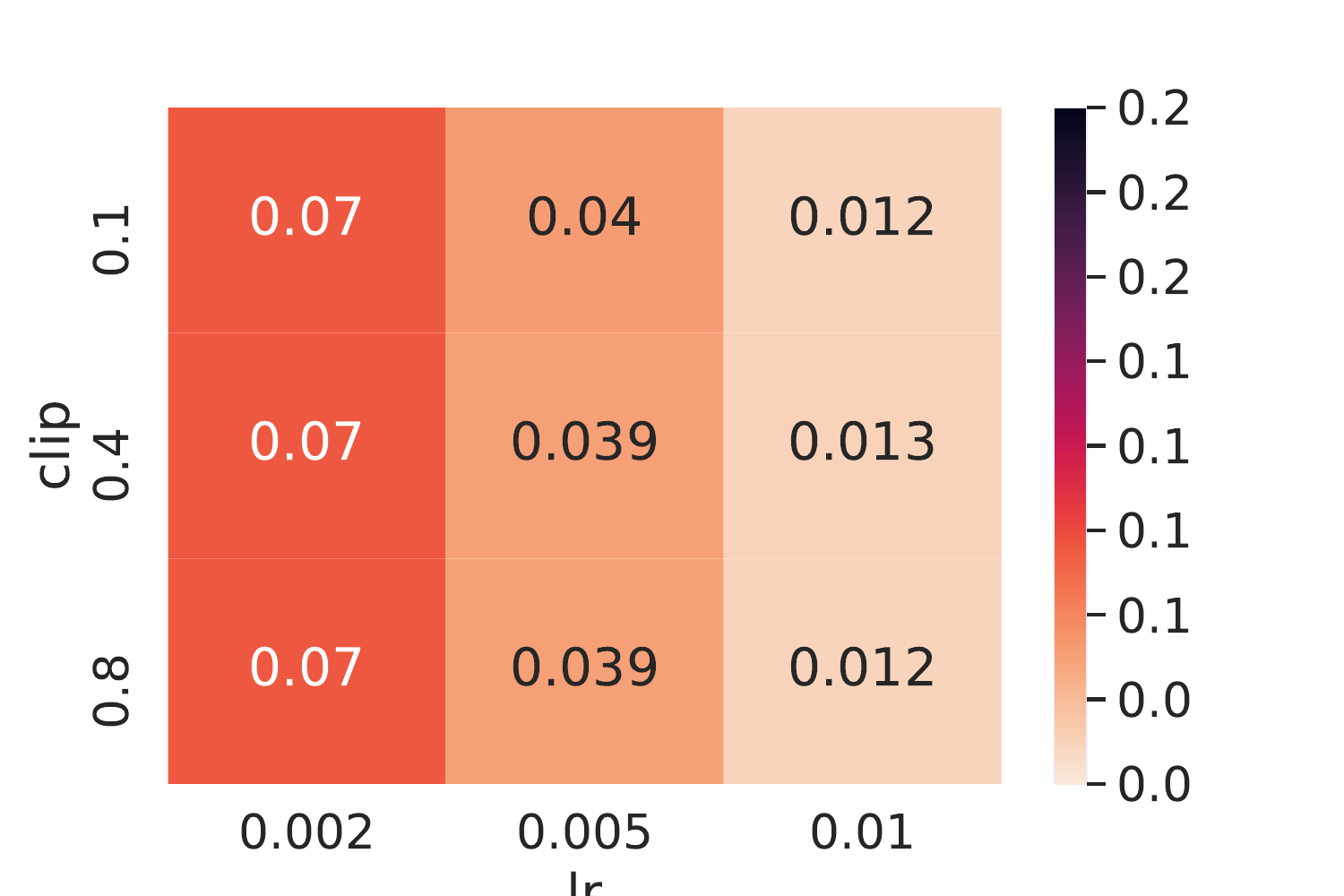}
     \footnotesize
     \caption{W.\@ Overlap for different clipping threshold \& LRs}
     \label{fig:s1:vocab:clip}
    \end{subfigure}
    \begin{subfigure}{0.23\textwidth}
     \includegraphics[width=\linewidth]{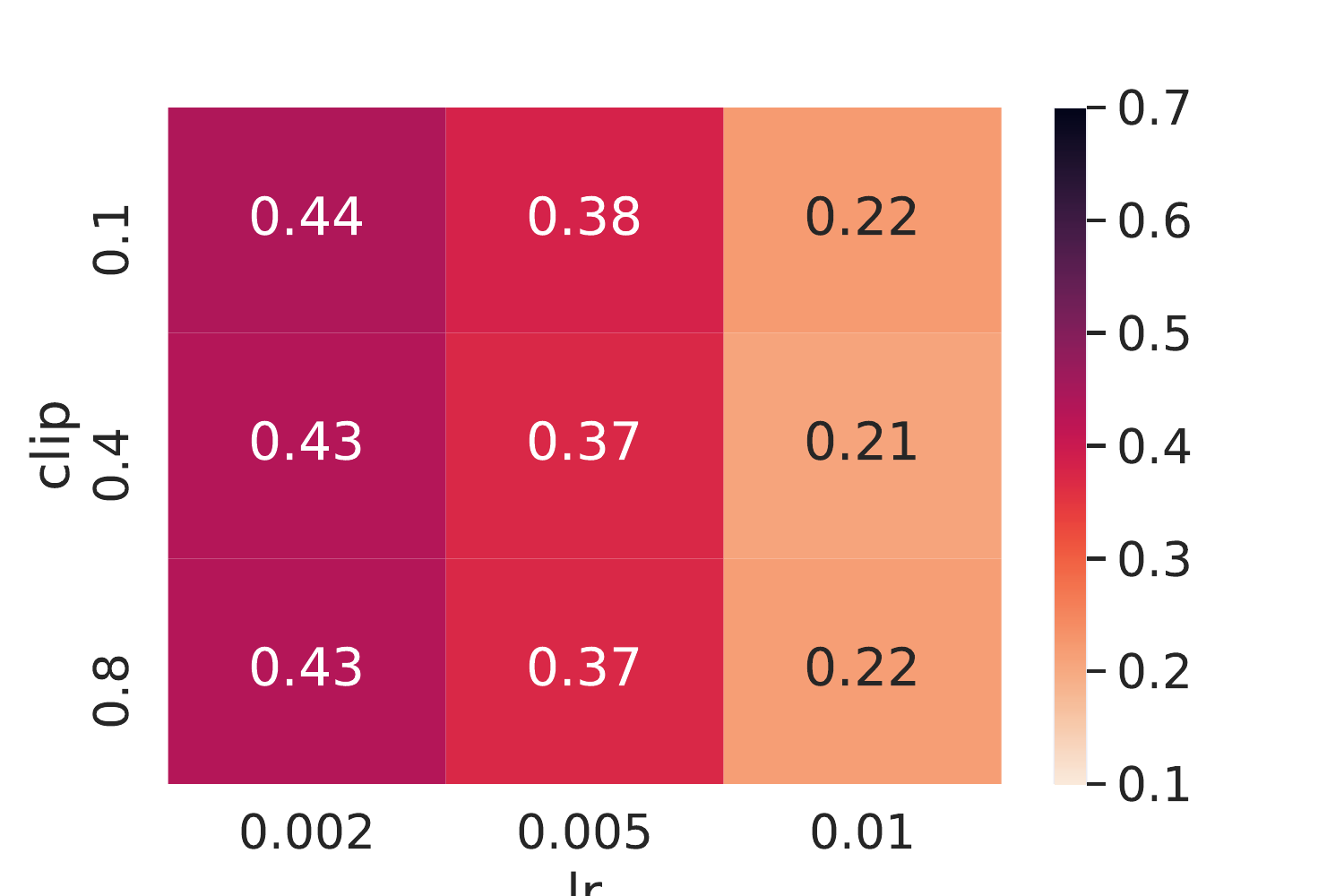}
    \footnotesize 
     \caption{F.\@ Overlap for different clipping threshold \& LRs}
     \label{fig:s1:node:clip}
    \end{subfigure}
    \begin{subfigure}{0.23\textwidth}
    \includegraphics[width=\linewidth]{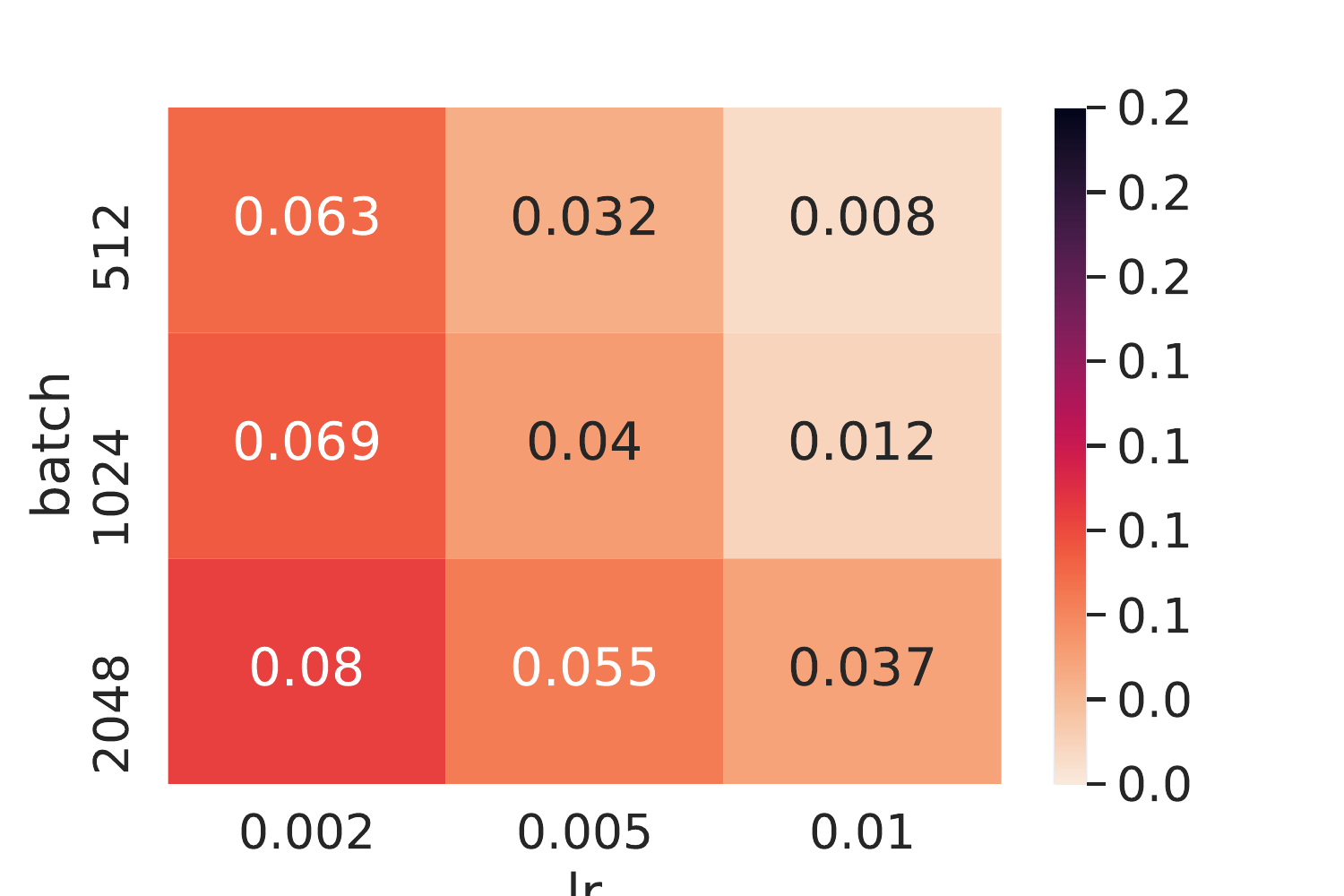}
     \footnotesize
     \caption{W.\@ Overlap for different batch sizes \& LRs}
     \label{fig:s1:vocab:bsize}
    \end{subfigure}
    \begin{subfigure}{0.23\textwidth}
    \includegraphics[width=\linewidth]{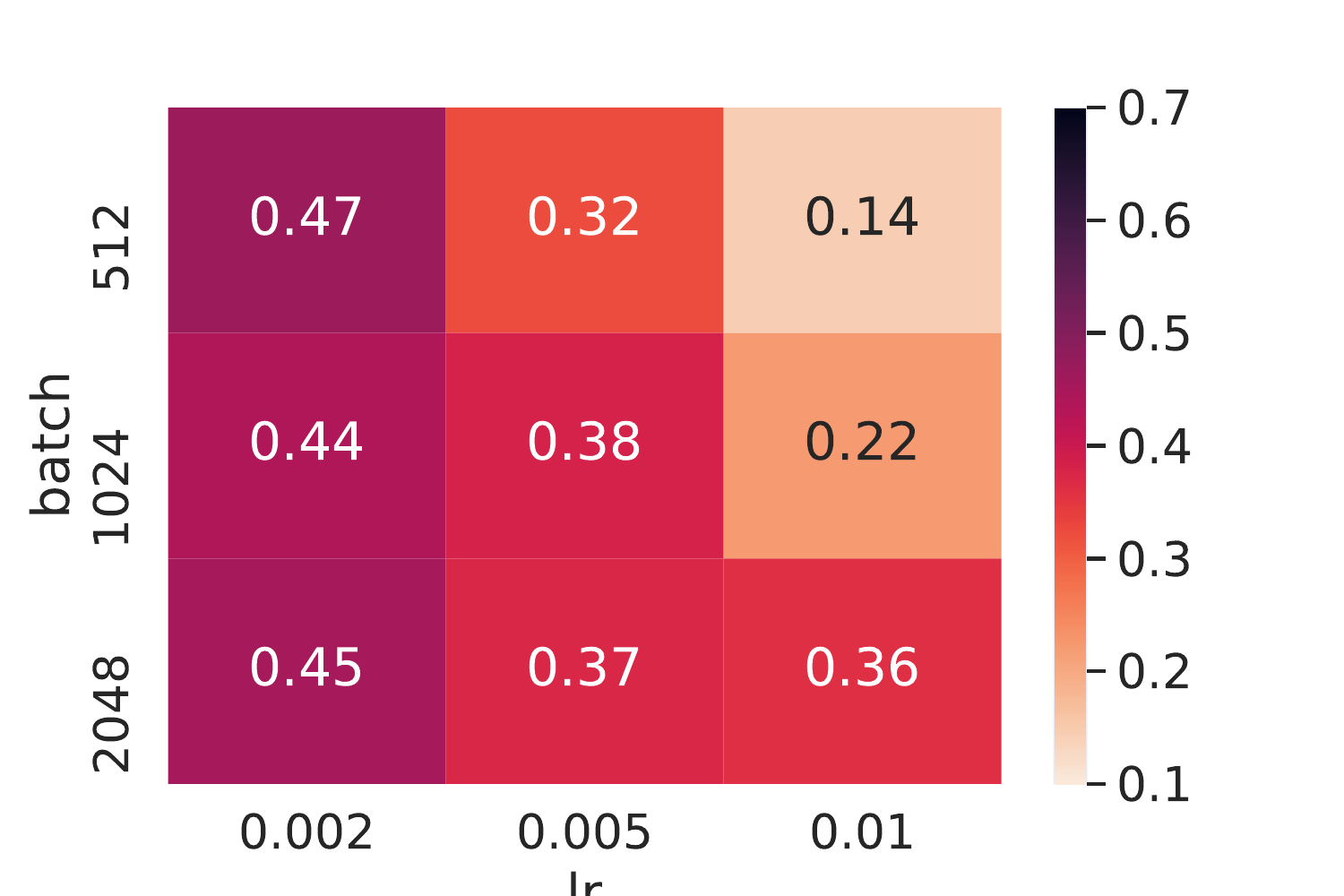}
     \footnotesize
     \caption{F.\@ Overlap for different batch sizes \& LRs}
     \label{fig:s1:node:bsize}
    \end{subfigure}

    \begin{subfigure}{0.23\textwidth}
     \includegraphics[width=\linewidth]{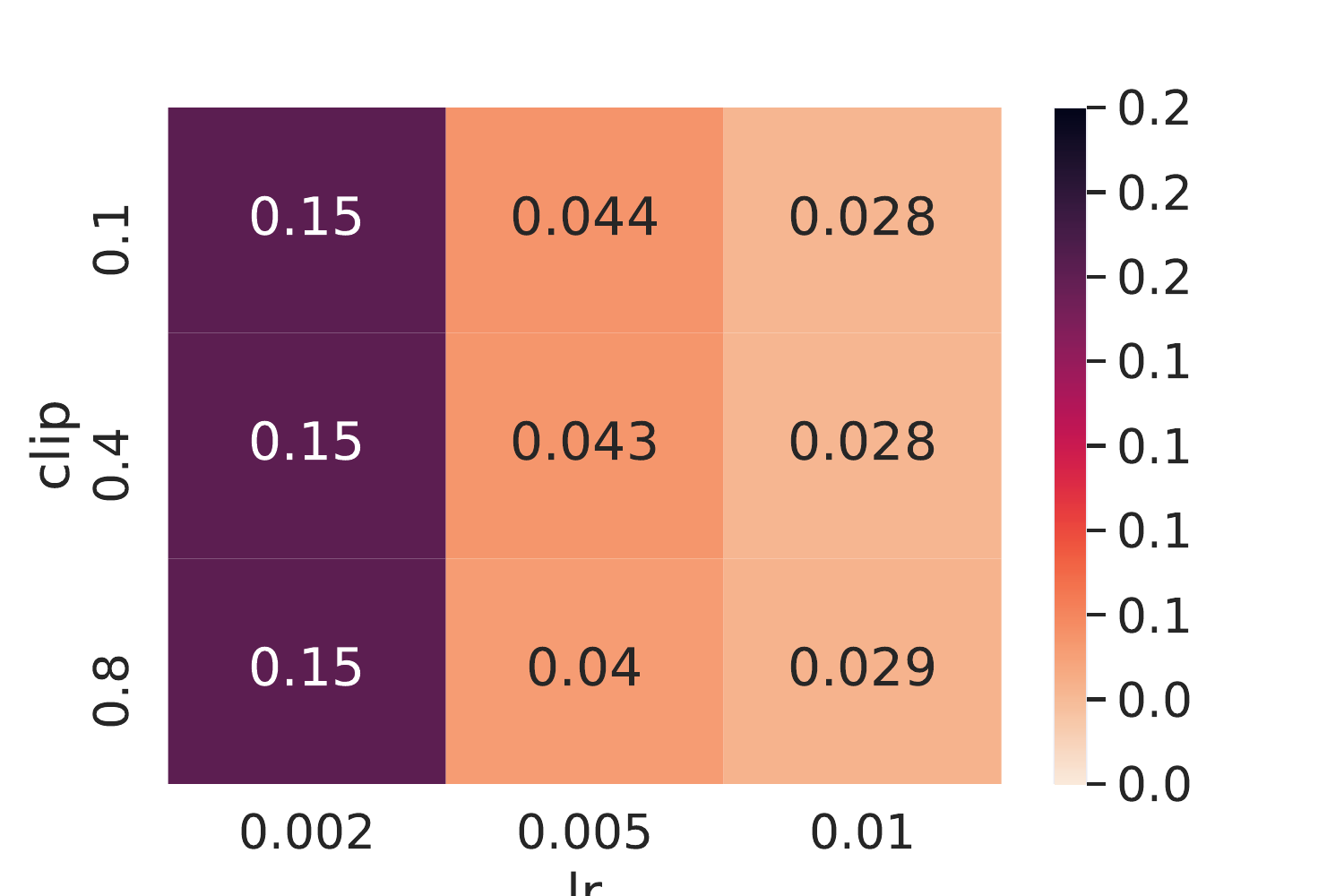}
     \footnotesize
     \caption{W.\@ Overlap for different clipping threshold  \& LRs}
     \label{fig:s2:vocab:clip}
    \end{subfigure}
    \begin{subfigure}{0.23\textwidth}
    \includegraphics[width=\linewidth]{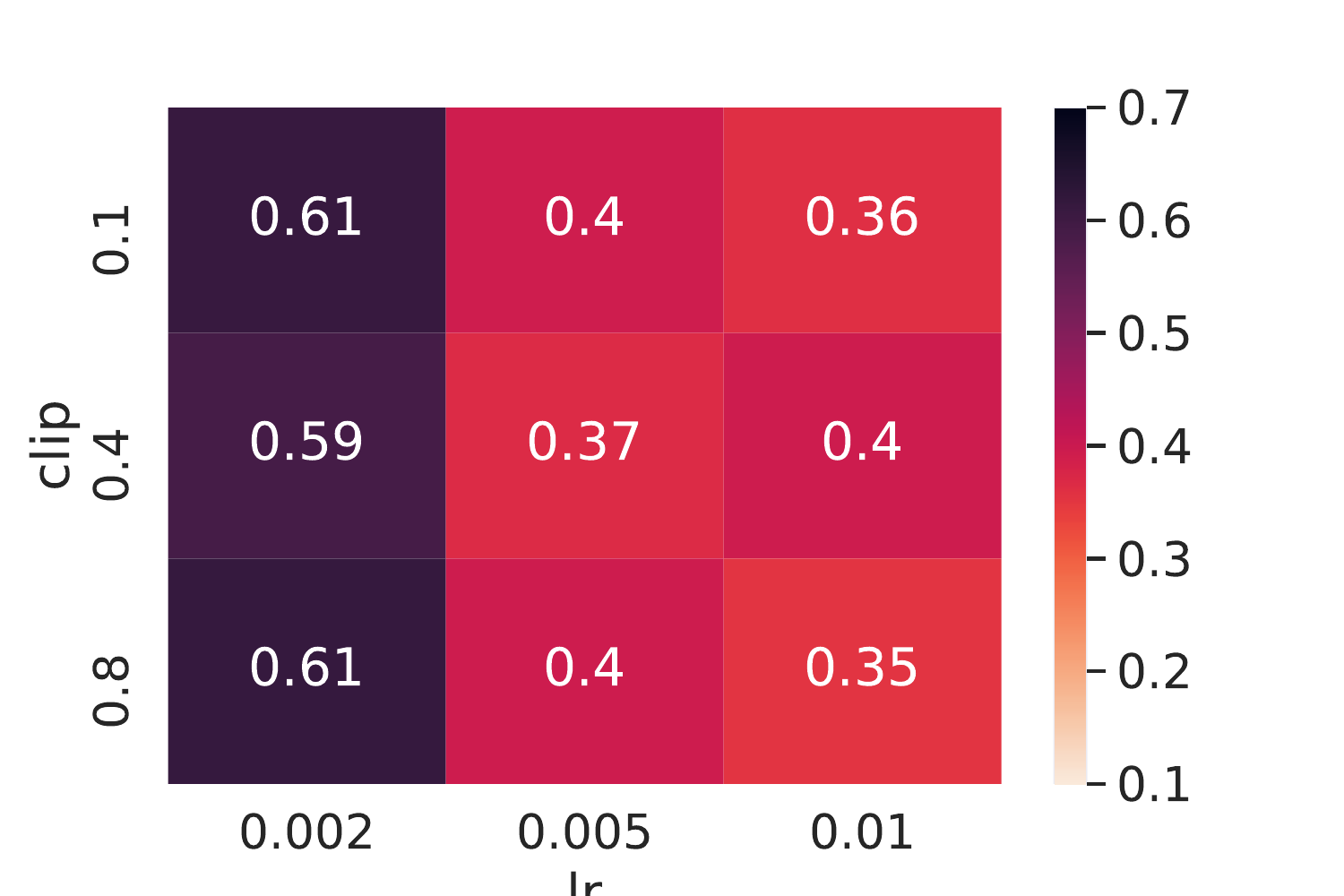}
    \footnotesize 
     \caption{F.\@ Overlap for different clipping threshold \& LRs}
     \label{fig:s2:node:clip}
    \end{subfigure}
    \begin{subfigure}{0.23\textwidth}
    \includegraphics[width=\linewidth]{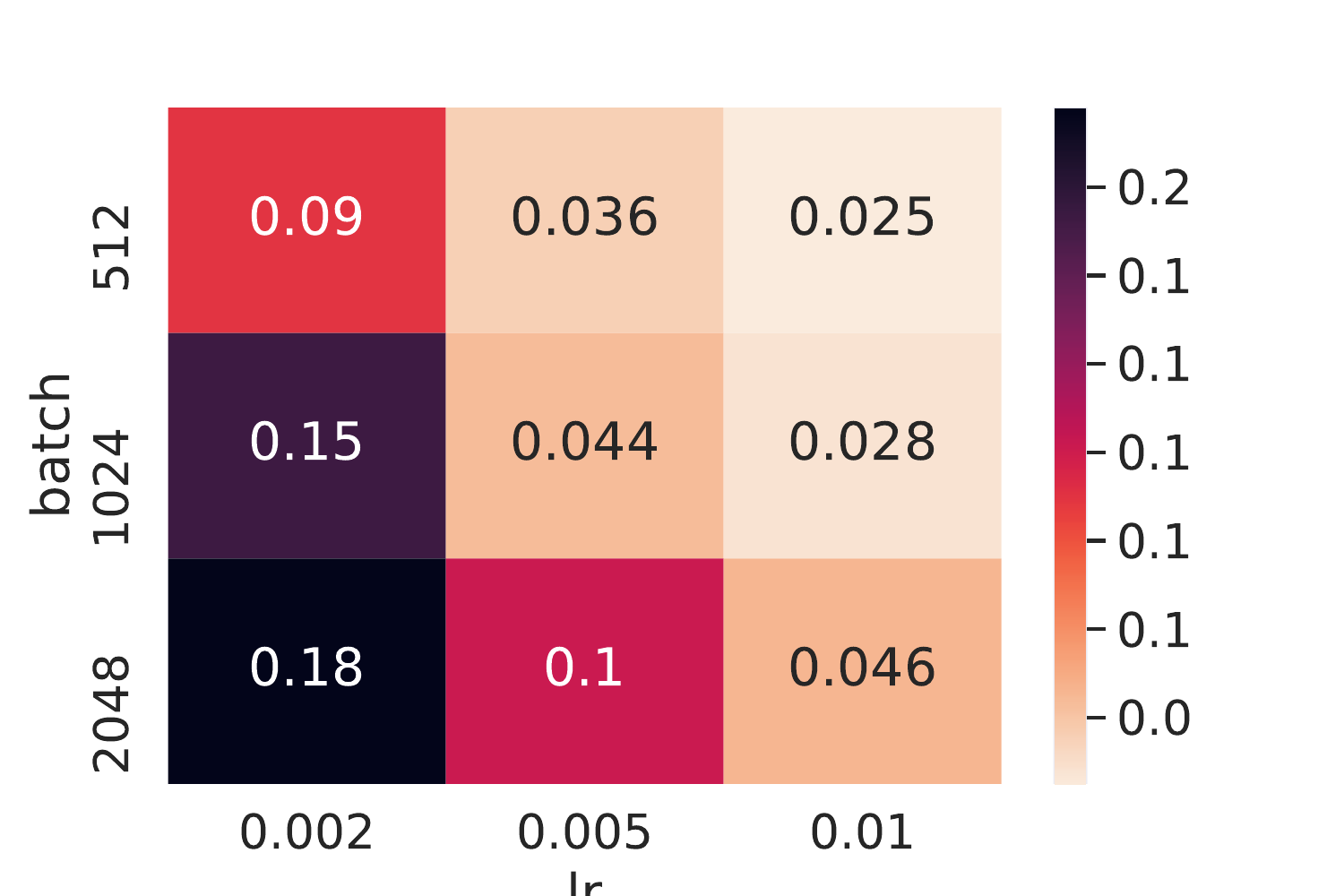}
     \footnotesize
     \caption{W.\@ Overlap for different batch sizes \& LRs}
     \label{fig:s2:vocab:bsize}
    \end{subfigure}
    \begin{subfigure}{0.23\textwidth}
    \includegraphics[width=\linewidth]{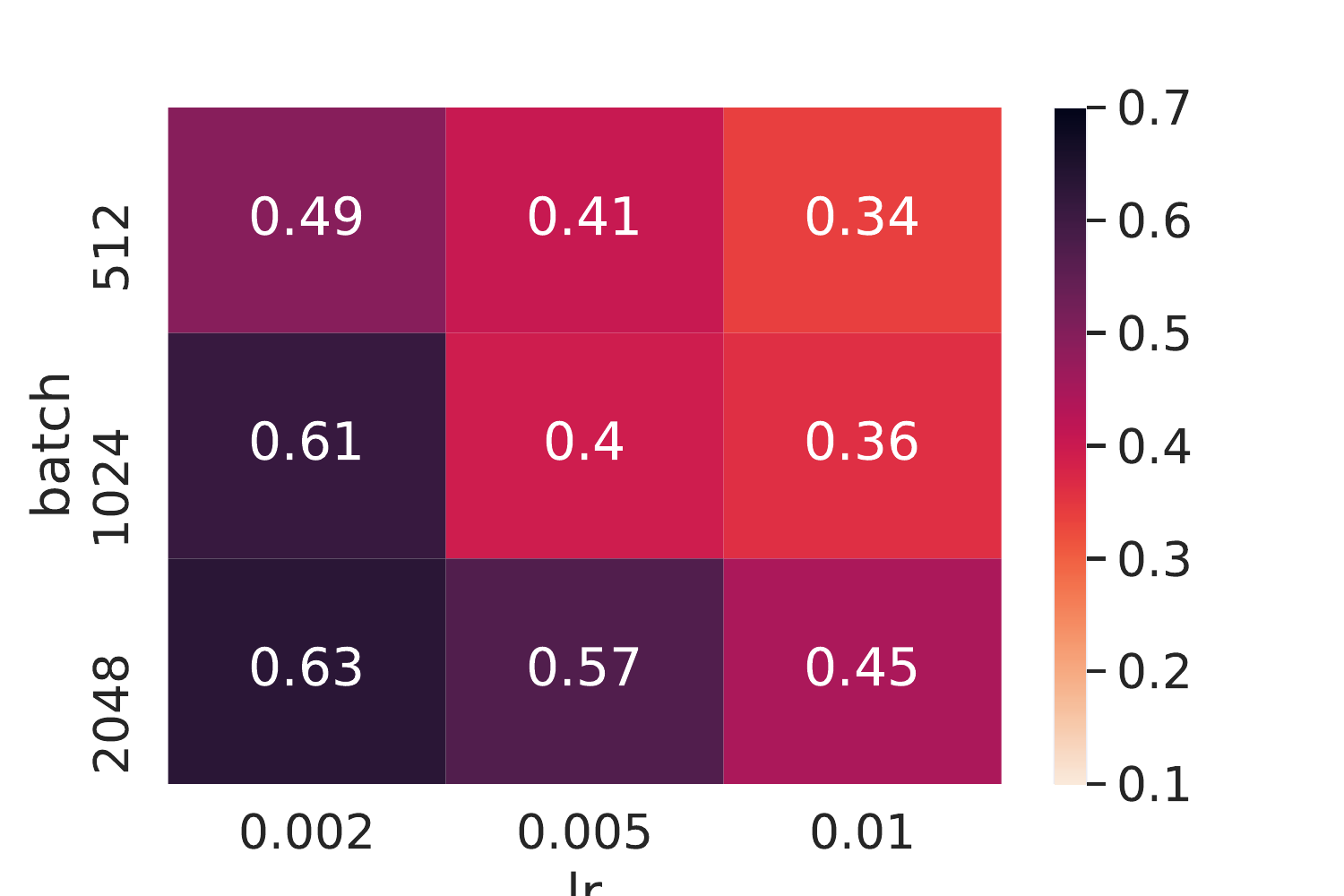}
     \footnotesize
     \caption{F.\@ Overlap for different batch sizes \& LRs}
     \label{fig:s2:node:bsize}
    \end{subfigure}
  \caption{Hyperparameter ablation studies for the 1-stage baseline (top) and 2-stage proposed method (bottom). We ablate the clipping threshold, learning rate and batch size. We report the vocab overlap and node overlap metrics, as language and parse metrics. } 
    \label{fig:ablation}
    \vspace{-2ex}
\end{figure*}

\section{Experimental Setup Details}\label{app:exp-details}

\subsection{DP Training Implementation}

We build our methods on top of the setup introduced by~\citet{li2021large}, and we use their repository~\url{https://github.com/lxuechen/private-transformers}.

\subsection{Software, Hardware, and Data Specifications}

 We use Opacus $0.15.0$, HuggingFace Transformers $4.10.3$, PyTorch $1.9.1$ with Cuda $10.2$,  and Python 3.8.8.
We run our experiments on an Azure ML Nvidia DGX-2 system, which has $16$ Tesla V100 GPUs with 512GB memory in total.
(We estimate that the experiments took an total of 4 weeks of GPU-hours.)
We use SMCalFlow 2.0, and TreeDST, which are both publicly available datasets.
For all the datasets and software used, we abide by the usage agreement.

\subsection{Training Hyperparameter Details}\label{app:hparams}

Based on the hyperparameter analysis shown in Section~\ref{app:hparam-analysis} below, we find the best clipping threshold to be $C=0.1$ in all experiments (SMCalFlow and TreeDST, and the learning rate to be \num{2e-3}.
We also set the privacy parameter $\delta=\num{8e-6}$, and $T=10$ epochs for the 2-stage setup. Below we list the dataset-specific parameters.

\paragraph{SMCalFlow.} For Table~\ref{tab:overview}, we use batch size of $2048$, by setting actual batch size to $32$ and gradient accumulation to $64$. For the rest of the tables we use batch size of $1024$ for the sake of speed, as we do gradient accumulation.

We use $T=10$ for $\epsilon=16$ and $T=7$ for $\epsilon$ values of $8$ and $3$ in the single stage scenario, and $T_1=2$ and $T_2=8$ in the 2-stage.
For the non-private experiments, $T=6$ for single stage and $T=10$ for two stage, where we split the epochs equally.
For the numbers in Tables~\ref{tab:modes} and~\ref{tab:shuffle} we set $T_1=3$ and $T_2=7$. 

\paragraph{TreeDST.} For Table~\ref{tab:overview}, we use batch size of $1024$, by setting actual batch size to $32$ and gradient accumulation to $32$. 
We use $T=10$ for all epsilon values in the single stage scenario, and $T_1=2$ and $T_2=8$ in the 2-stage. We set the learning rate to $0.002$.
For the non-private experiments, $T=6$ for 1-stage and $T=20$ for 2-stage, where we split the epochs equally. We set the learning rate to $0.001$.

\paragraph{High-Resource Parser Hyperparameters.}
 We use the parser architecture from~\citet{zhou2022online}, where we train the model with no context (as in we train on single utterances and not conversations), and we only use the human utterances and not the agent’s. We train two parsers, one on SMCalFlow and one on TreeDST, for their corresponding evaluations. All of our parsers are based on the Transformer architecture, adapted to the graph action sequence. Implementation details for the parser are provided in Appendix E of~\citet{zhou2022online}. We trained this model with learning rate of $0.001$ for $50$ epochs. 

\paragraph{End-to-end (Parser Improvement) Experiment Hyperparameters.}
For the experiments in Section~\ref{sec:end-to-end}, for training the DP models we use the same hyperparameters as before. For training the low-resource parsers, we find that for the one missing the Weather function the best set learning rate and epoch numbers are  $0.003$ and 150 epochs.
For event on date we use the same learning rate, but train for 180 epochs.

\paragraph{Choosing the best setup to report.} To select the best hyperparameter setup for the baselines and our method, we relied on MAUVE and Function overlap metrics, as in chose the setup which had a higher combination of these two values (in almost all cases if one was highest the other was also highest). However, we did discard some setups with high GPT-2 loss (higher than $2.0$ on a GPT-2 fine-tuned on SMcalflow 2.0), as we observe that MAUVE and Function overlap tend to be very high on some grammatically incorrect sentences with diverse vocabulary. Therefore, we use GPT-2 loss as an auxiliary metric to help us sift. 

\section{Additional Experimental Results}\label{app:additional}

\subsection{Hyperparameter Sensitivity Analysis}\label{app:hparam-analysis}

In this section we compare different hyperparameter settings in the generation process, and analyze their effect on the quality of the synthesized text. First, we vary the privacy budget split between the first and second stages, in the 2-stage setup. Then, we vary the training hyperparameters such as learning rate, clipping threshold, and batch size, to measure the sensitivity of the results. We use the test set portion of the dataset for this.

\subsubsection{Training Epoch Split}
How the number of training epochs is split between the two stages (i.e., how the privacy budget is split) has a significant impact on the quality of the synthesized text. 
Figure~\ref{fig:budget} shows the results for this experiment, where we experiment with all $9$ possible ways of splitting $10$ total training epochs between the two stages. A row label such as ``$9$--$1$'' denotes a model trained for $9$ epochs on stage 1 and $1$ epoch on stage 2.
The more epochs spent on a stage, the more the privacy expenditure of that stage.

Based on the $4$ measured metrics, we can see that the setup $2$--$8$ is the best one. We also see that as we go from the bottom of the graph to the top (i.e. as we spend less epochs on training the parse generation model and more epochs on training the parse2utterance) the overall quality improves.  
However, we also see that this trend breaks when we go from $2$--$8$ to $1$--$9$, where the metrics plummet, which is probably due to $1$ training epoch being too small for the parse generation model. 

\subsubsection{Clipping, Batch size and Learning Rate}

Finally, we run a hyperparameter sensitivity analysis on the gradient clipping threshold (for the DP optimization which requires gradient clipping and addition of noise, see Section~\ref{sec:background}), batch size and learning rate. 
Figure~\ref{fig:ablation} shows the results for the 1-stage and 2-stage techniques, on the top and bottom rows of the figure, respectively.
Here, we only look at the token type overlap and function overlap metrics, for simplicity. The first two graphs in each row are controlled for batch size  (set to $1024$ and change clipping and learning rate), whereas the last two graphs are controlled for clipping threshold (set to $0.1$) and change batch size and learning rate. 
We can see that for both the baseline and the proposed method, the best set of parameters is $0.1$, $0.002$ and $2048$ for clipping, learning rate and batch size, respectively.

\subsection{Disrupting the Correlation Between Parse Trees and Utterances} \label{app:shuffle}
For this experiment we take the parse trees and utterances in the dataset and randomly pair them up,
so that the utterance in each pair has no relationship to the parse tree.  In this setting, we expect the parse2utterance model to ignore the uninformative parse or, by fitting to spurious correlations, use it as a source of randomness.  

Table~\ref{tab:shuffle} shows the results for this experiment, both with and without DP.
We can see that the 2-stage shuffle model performs worse than the baseline on the parse-related metrics, which supports the hypothesis that the structure in the parse trees and the correlation to utterances are important for these metrics.  That is, the benefit of the original 2-stage model does not arise simply from the fact that it has 2-stages or more parameters.
We also see the same pattern for the word-type (vocabulary) overlap metric, where once shuffled the model fails to capture the diversity in the vocabulary. However, the baseline exhibits higher MAUVE than the shuffled model.
Manual investigation revealed that the shuffled model's generations are grammatically correct but are more repetitive across utterances,  compared to other models, and MAUVE fails to sufficiently penalize these issues.

\begin{table*}[]
    \centering
    \vspace{-2ex}
    \begin{adjustbox}{width=0.7\linewidth, center}
     \newcolumntype{L}{>{\RaggedLeft\arraybackslash}p{0.06\linewidth}} 
  \newcolumntype{O}{>{\RaggedLeft\arraybackslash}m{0.07\linewidth}} 
  \newcolumntype{D}{>{\arraybackslash}m{0.15\linewidth}} 
  \newcolumntype{R}{>{\arraybackslash}m{0.29\linewidth}} 
\begin{tabular}{@{}clcccccSSSS@{}}

	\toprule
	& {\multirow{2}{*}{}} &  \multicolumn{2}{c}{{Language Metrics}}&{}& \multicolumn{2}{c}{{Parse Metrics
 }}   \\
	\cmidrule{3-4} \cmidrule{6-7}
	&  Method &  {W.\@ Overlap $\uparrow$}  &{MAUVE $\uparrow$}&{} & {Dist.\@ $\downarrow$} & {F.\@ Overlap $\uparrow$}  \\
    \midrule
    \multirow{1}{*}{\STAB{}} 
    &  Ground Truth & 1.0	& 1.0 &	&0.00296&	0.82	\\
    \midrule[0.1pt] 
   \multirow{3}{*}{\STAB{No DP}} 
    &  1-stage                  &0.087$\pm$0.005&0.334$\pm$0.056&&0.258$\pm$0.034&0.487$\pm$0.004	\\
    & 2-stage                   &0.236$\pm$0.012&0.632$\pm$0.005&&0.085$\pm$0.009&0.797$\pm$0.006  \\
    & 2-stage-shuffle           &0.049$\pm$0.009&0.560$\pm$0.031&&0.354$\pm$0.070&0.391$\pm$0.019  \\
           \midrule[0.1pt]
    \multirow{3}{*}{\STAB{$\epsilon=3$}}
    &  1-stage              &0.086$\pm$0.006&0.160$\pm$0.072&&0.170$\pm$0.023&0.475$\pm$0.018	\\
    & 2-stage               &0.169$\pm$0.007&0.404$\pm$0.063&&0.079$\pm$0.010&0.639$\pm$0.005  \\
    & 2-stage-shuffle       &0.032$\pm$0.004&0.401$\pm$0.063&&0.453$\pm$0.024&0.232$\pm$0.010  \\

	\bottomrule
\end{tabular}

    \end{adjustbox}
    \caption{The effect that disrupting the correlation between parse trees and utterances (shuffling parse trees and utterances) has on the performance of the 1-stage and 2-stage models.  The goal here is to confirm that the superiority of the 2-stage model is due to its exploiting the correlations between parses and utterances. The numbers reported are presented in the format of mean $\pm\sigma$, over three runs with three different seeds.}
    \vspace{-2ex}
        \label{tab:shuffle}
\end{table*}

\subsection{Metrics for the Samples Used in the  Parser improvement Experiment}\label{app:e2ebreak}
Table~\ref{tab:end-met-break} shows the detailed language and parse metrics for the checkpoints used for augmenting the parser in Section~\ref{sec:end-to-end}. These results are obtained for $\epsilon=3$, and for the 2-stage method they are overall worse than the numbers shown in the main results, Table~\ref{tab:overview}, for the same $\epsilon$ value. The reason behind this discrepancy is that for the parser improvement experiment we annotate the training samples using the low-resource parser which is much less accurate that the ground truth parse trees used in Table~\ref{tab:overview}. However, we see that for the 1-stage baseline the overlap metrics are now much higher. The reason behind this discrepancy is that in Table~\ref{tab:end-met-break} we report results over the 90k samples, whereas the results in the body of the paper are over 13k samples, and the more we generate the more vocabulary/functions get covered.

\begin{table}[]
    \centering
    \vspace{-2ex}
    \begin{adjustbox}{width=0.99\linewidth, center}
     \newcolumntype{L}{>{\RaggedLeft\arraybackslash}p{0.06\linewidth}} 
  \newcolumntype{O}{>{\RaggedLeft\arraybackslash}m{0.07\linewidth}} 
  \newcolumntype{D}{>{\arraybackslash}m{0.15\linewidth}} 
  \newcolumntype{R}{>{\arraybackslash}m{0.29\linewidth}} 
\begin{tabular}{@{}clcccccSSSS@{}}

	\toprule
	& {\multirow{2}{*}{}} &  \multicolumn{2}{c}{{Language Metrics}}&{}& \multicolumn{2}{c}{{Parse Metrics
 }}   \\
	\cmidrule{3-4} \cmidrule{6-7}
	&  Method &  {W.\@ Overlap $\uparrow$}  &{MAUVE $\uparrow$}&{} & {Dist.\@ $\downarrow$} & {F.\@ Overlap $\uparrow$}  \\
    \midrule
    \multirow{1}{*}{\STAB{}} 
    &  Ground Truth & 1.0	& 1.0 &	&0.00296&	0.82	\\
    \midrule[0.1pt] 
   \multirow{2}{*}{\STAB{\textcolor{black}{Wea.}}} 
    &  1-stage       &0.163	&0.282&	&0.142	&0.656	\\
    & 2-stage        &0.294	&0.460&	&0.088	&0.729  \\
\midrule[0.1pt]
   \multirow{2}{*}{\STAB{\textcolor{black}{EoD}}} 
    &  1-stage      &0.115	&0.137&	&0.221	&0.536	\\
    & 2-stage       &0.313	&0.471&	&0.078	&0.758  \\
	\bottomrule
\end{tabular}
    \end{adjustbox}
    \caption{The language and parse metrics for the generations used in the end-to-end, parser improvment experiments (Table~\ref{tab:end-to-end-main}). Wea.\@ shows the results for dropping the Weather functions and EoD shows experiments for dropping EventOnDate function. We do not report $\sigma$ here since we use only one checkpoint's generations for augmentation.}
        \label{tab:end-met-break}
    \vspace{-2ex}
\end{table}

\subsection{Function Distribution Coverage Break-down Results}\label{app:coverage}

To provide a better, more detailed depiction of how well each method matches the distribution of the functions to that of the ground truth, we report the overlap between the top-10, 25, 50 and 100 most common functions in the ground truth, with those for each generation method. We report those results in Table~\ref{tab:break-coverage}.

\begin{table*}[]
    \centering
   \begin{adjustbox}{width=0.82\textwidth, center}
    
 \newcolumntype{L}{>{\RaggedLeft\arraybackslash}p{0.06\linewidth}} 
  \newcolumntype{O}{>{\RaggedLeft\arraybackslash}m{0.07\linewidth}} 
  \newcolumntype{D}{>{\arraybackslash}m{0.15\linewidth}} 
  \newcolumntype{R}{>{\arraybackslash}m{0.29\linewidth}} 
\begin{tabular}{@{}clcccccccccccccccc@{}}
	\toprule
	&  {\multirow{3}{*}{}} &\multicolumn{4}{c}{SMCalFlow} &{} &\multicolumn{4}{c}{TreeDST}\\
 	\cmidrule{3-6} \cmidrule{8-11}
	&  Method &  {Top-10}  &{Top-25}&{Top-50} & {Top-100}& &{Top-10}  &{Top-25}&{Top-50} & {Top-100} \\
    \midrule
           \midrule[0.1pt]
    \multirow{2}{*}{\STAB{$\epsilon=16$}}
    &  1-stage    &0.40	&0.60	&0.80	&0.84	&&1.00	&0.68	&0.74	&0.81    \\
    & 2-stage     &0.60	&0.76	&0.92	&0.90	&&1.00	&0.80	&0.88	&0.87    \\
           \midrule[0.1pt]
    \multirow{2}{*}{\STAB{$\epsilon=8$}}
    &  1-stage          &0.50	&0.64	&0.80	&0.84	&&1.00	&0.72	&0.78&	0.81 	\\
    & 2-stage           &0.60	&0.80	&0.92	&0.90	&&1.00	&0.80	&0.88&	0.88   \\
           \midrule[0.1pt]
    \multirow{2}{*}{\STAB{$\epsilon=3$}}
    &  1-stage          &0.60	&0.64	&0.82	&0.84	&&1.00	&0.72	&0.82	&0.81  	\\
    & 2-stage           &0.60	&0.80	&0.94	&0.89	&&1.00	&0.80	&0.88	&0.87    \\
	\bottomrule
\end{tabular}

    \end{adjustbox}
    \caption{Breakdown of the function type distribution/coverage for different methods (Table~\ref{tab:overview}). We show the overlap of the top-k function types between ground truth and the generations.}
        \label{tab:break-coverage}
\end{table*}

\subsection{Results For Larger Models}\label{app:large}
The results discussed in the body of the paper are all reported on pre-trained GPT2-small. To further explore with other models, we present results on GPT2-Large, on the SMCalFlow dataset and present the results in Table~\ref{tab:large}. We do not observe improvements over GPT2-small in the synthesized data’s quality.  We speculate this is due to insufficient hyperparameter searches on the larger model, or the small size of our training dataset. 

\begin{table}[]
    \centering
    \vspace{-2ex}
    \begin{adjustbox}{width=0.99\linewidth, center}
     \newcolumntype{L}{>{\RaggedLeft\arraybackslash}p{0.06\linewidth}} 
  \newcolumntype{O}{>{\RaggedLeft\arraybackslash}m{0.07\linewidth}} 
  \newcolumntype{D}{>{\arraybackslash}m{0.15\linewidth}} 
  \newcolumntype{R}{>{\arraybackslash}m{0.29\linewidth}} 
\begin{tabular}{@{}clcccccSSSS@{}}

	\toprule
	& {\multirow{2}{*}{}} &  \multicolumn{2}{c}{{Language Metrics}}&{}& \multicolumn{2}{c}{{Parse Metrics
 }}   \\
	\cmidrule{3-4} \cmidrule{6-7}
	&  Method &  {W.\@ Overlap $\uparrow$}  &{MAUVE $\uparrow$}&{} & {Dist.\@ $\downarrow$} & {F.\@ Overlap $\uparrow$}  \\
    \midrule

   \multirow{2}{*}{\STAB{\textcolor{black}{NoDP}}} 
    &  1-stage       &0.049$\pm$0.007&0.166$\pm$0.032&&0.375$\pm$0.121&0.345$\pm$0.019	\\
    & 2-stage       &0.230$\pm$0.001&0.608$\pm$0.037&&0.088$\pm$0.003&0.781$\pm$0.013 \\
\midrule[0.1pt]
   \multirow{2}{*}{\STAB{\textcolor{black}{$\epsilon=8$}}} 
    &  1-stage      &0.104$\pm$0.006&0.136$\pm$0.019&&0.172$\pm$0.010&0.485$\pm$0.021\\
    & 2-stage      &0.187$\pm$0.007&0.476$\pm$0.028&&0.108$\pm$0.008&0.674$\pm$0.015 \\
	\bottomrule
\end{tabular}
    \end{adjustbox}
    \caption{Results for GPT2-Large}
        \label{tab:large}
    \vspace{-2ex}
\end{table}
\subsection{Effect of Data-augmentation for Adding  New Functionality on Existing Functions}\label{app:new-column}

In this section we want to see whether our data augmentation we implement to add new functionality ends up hurting the performance of the existing functions. Table~\ref{tab:end-to-end-app} shows this. This table corresponds to Table~\ref{tab:end-to-end-main} in the main body of the paper, however it presents results for every functionality, except the augmented one.   

\begin{table*}[]
    \centering
    \vspace{-2ex}
   \begin{adjustbox}{width=0.99\linewidth, center}

 \newcolumntype{L}{>{\RaggedLeft\arraybackslash}p{0.06\linewidth}} 
  \newcolumntype{O}{>{\RaggedLeft\arraybackslash}m{0.07\linewidth}} 
  \newcolumntype{D}{>{\arraybackslash}m{0.15\linewidth}} 
  \newcolumntype{R}{>{\arraybackslash}m{0.29\linewidth}} 
\begin{tabular}{@{}clccccccccc@{}}

	\toprule
	& {\multirow{2}{*}{}} &  \multicolumn{3}{c}{{Missing Weather function  breakdown
 }}&{}& \multicolumn{3}{c}{{Average over all function types}}   \\
	\cmidrule{3-5} \cmidrule{7-9}
	&  Method &  Anonymized String Match&	API Precision	&API Recall&{} & Anonymized String Match&	API Precision	& API Recall \\
    \midrule
    \multirow{1}{*}{\STAB{}} 
   \multirow{3}{*}{\STAB{\footnotesize{Weather}}} 
&Non-augmented	&0.0$\pm$0.0&1.9$\pm$0.7&2.1$\pm$0.9    && 60.9$\pm$1.1&65.7$\pm$1.2&65.6$\pm$1.3\\					    
&  1-stage      &37.7$\pm$2.6&42.1$\pm$3.1&42.1$\pm$3.1 && 63.4$\pm$3.1&74.9$\pm$5.0&75.0$\pm$4.9	 	    \\
& 2-stage       &43.7$\pm$0.7&50.3$\pm$0.9&50.3$\pm$0.9 && 66.5$\pm$0.3&77.6$\pm$5.0&77.5$\pm$4.9	      \\

	\bottomrule
\end{tabular}

    \end{adjustbox}
    \caption{Low-resource semantic parser augmentation experiment results. The Weather and EoD rows determine the composition of $\Dpub$ (Section~\ref{sec:end-to-end}). The ``Missing (Weather)'' columns report the metrics over only the functionality that was missing from the public data, but present in $\Duser$. The ``All But Missing'' columns report metric over all \textbf{other} function-types, to see if we observe degradation from the augmentation.}
         \label{tab:end-to-end-app}
    \vspace{-2ex}
\end{table*}

\subsection{Examples of Generated Utterances and Parse-trees}
Table~\ref{tab:generations} shows some examples of generated utterances for the 1-stage baseline, and the 2-stage proposed method. 
Table~\ref{tab:trees_gens} shows pairs of generated trees and then utterances that are conditioned on those trees, for the second stage model. 
We observe some syntactically invalid generated parse trees from the parse generation model in the 2-stage setup. Nonetheless, the second stage model can still generate coherent text from such trees, and benefit from them, as we see the 2-stage model captures the distribution better.

\section{DP and Privacy accounting}\label{app:dp-sgd}

\subsection{Training via DP-SGD}\label{sec:background}
To train a neural network with differential privacy, the most widely used algorithm is the  DP variant of stochastic gradient descent (DP-SGD)~\citep{abadi2016deep}.
DP-SGD resembles ordinary SGD, but at each gradient update step (derived from a minibatch of training examples), it  first clips the per-example gradient by its norm, then obfuscates the gradient by adding Gaussian noise.
Intuitively, 
this limits the contribution that a single example makes to the final model parameters.

Clipping the gradient to a maximum norm of $C$ (a hyperparameter) is done as follows:
\begin{align}
\tilde g \leftarrow \sum_{x \in B}{\sf clip}\left(\nabla \ell(x)\right)
\end{align}
where $B$ is the current minibatch and ${\sf clip}(v)=v\cdot\min\left\{1,\frac{C}{\|v\|_2}\right\}$. 
The clipped gradient is small enough for us to obfuscate it (without further changing its mean) by adding Gaussian noise with mean 0 and standard deviation $\sigma C$:
\begin{align}
g \leftarrow \frac{\tilde g+\mathcal{N}\left(0, \sigma^2 C^2\right)}{|B|}
\end{align}

The privacy expenditure $(\epsilon, \delta)$ is a function of $C$, $\sigma$, $|B|$, $|\cD|$, and the total number of epochs $T$ (which controls the total number of gradient updates). It is determined based on the R\'enyi DP~\cite{mironov2017renyi} privacy accounting method. In practice, following prior work, we fine-tune our models using DP-Adam~\cite{abadi2016deep,li2021large}.

\paragraph{Post-processing property.} The post-processing property of DP \cite{DworkKMMN06} ensures that if an algorithm $\mathcal{A}$ satisfies
$(\epsilon, \delta)$-DP, then so does  $F\circ{A}$ for any function $F$, which means that we can run as many inferences (i.e., take as many samples) as we want from the DP-trained models, without changing the privacy expenditure.

\begin{table*}[]
    \centering
    \begin{adjustbox}{width=0.98\linewidth, center}
     \newcolumntype{L}{>{\RaggedLeft\arraybackslash}p{0.06\linewidth}} 
  \newcolumntype{O}{>{\RaggedLeft\arraybackslash}m{0.07\linewidth}} 
  \newcolumntype{D}{>{\arraybackslash}m{0.15\linewidth}} 
  \newcolumntype{R}{>{\arraybackslash}m{0.47\linewidth}} 
\begin{tabular}{@{}l@{\hskip 2mm}p{10.5cm}p{10.5cm}@{}}
	\toprule
	& {Single-stage baseline} & { Two-stage proposed method} \\

    \midrule

   \multirow{5}{*}{\STAB{No DP}}
&What is the weather like in San Francisco? 
&what will the weather be like tomorrow in texas?
 \\
&Can you find any meetings scheduled between 2 and 5 pm next Friday? 
&Do I have any more meetings scheduled in the next two weeks?
\\
&I need to cancel all my meetings for the rest of the week. 
& Cancel the marketing meeting.
\\
& also i need to book a hotel room for the end of the day on friday 
& Change my snowboarding event next month to end at 2 pm.
\\
& invite all my relatives to the baby shower 
& Who has declined the invite for my vacation this month?
\\

    \midrule
    \multirow{5}{*}{\STAB{$\epsilon$=3}}
&   what is the weather going to be like for this weekend?
&   What is the weather in Seattle today?
\\
& Can you tell me what events I have scheduled for next week? 
& Can you tell me if I have any appointments scheduled for next week?
\\
&   cancel the meeting with Kim and her team. 
&  I need to cancel my meeting with my supervisor.
\\
&   all events are supposed to end at 1 pm instead of 2 pm. 
&   Can you make it end at 2 pm?
\\
&   who is attending the meeting?  
&   Can you schedule a meeting with other attendees today?
\\
	\bottomrule
\end{tabular}

    \end{adjustbox}
         \caption{Sample generations from the 1-stage baseline and the 2-stage proposed method, with and without differential privacy. The sentences that are put in one row were generated completely independently and have no particular correspondence; we only tried to group sentences based on similarity. The hyperparameters here are those yoused to generate Table~\ref{tab:overview}.}
        \label{tab:generations}
\end{table*}

\begin{table*}[]
    \centering
    \begin{adjustbox}{width=0.98\linewidth, center}
     \newcolumntype{L}{>{\RaggedLeft\arraybackslash}p{0.06\linewidth}} 
  \newcolumntype{O}{>{\RaggedLeft\arraybackslash}m{0.07\linewidth}} 
  \newcolumntype{D}{>{\arraybackslash}m{0.15\linewidth}} 
  \newcolumntype{R}{>{\arraybackslash}m{0.47\linewidth}} 
\begin{tabular}{@{}l@{\hskip 2mm}p{10.5cm}p{10.5cm}@{}}
	\toprule
	& {Parse trees ($y$)} & {Utterances ($x$)} \\

    \midrule

   \multirow{5}{*}{\STAB{No DP}}
& \begin{lstlisting}
    move (Yield (DeleteCommitEventWrapper (DeletePreflightEventWrapper (Event.id (singleton (QueryEventResponse.results (FindEventWrapperWithDefaults (Event.subject\_? (?~= "marketing meeting")))))))))
\end{lstlisting}
& Cancel the marketing meeting
 \\
& 
\begin{lstlisting}
wasYield (AttendeesWithResponse (Event.attendees (singleton (QueryEventResponse.results (FindEventWrapperWithDefaults (EventDuringRange (Event.subject_? (?~= "vacation"))) (FullMonthofMonth (Date.month (Today)))))) (ResponseStatusType.Declined))) 
\end{lstlisting}
& Who has declined the invite for my vacation this month?
\\

    \midrule
    \multirow{5}{*}{\STAB{$\epsilon$=3}}
& \begin{lstlisting}   
at Aerospace (Yield (WeatherQueryApi (AtPlace (FindPlace (LocationKeyphrase.apply "Seattle")))))))) 
\end{lstlisting}
&  What is the weather in Seattle
\\
& 
\begin{lstlisting}
(Yield (Execute (ReviseConstraint (refer (^(Dynamic) roleConstraint (Path.apply "output"))) ((Event) ConstraintTypeIntension) (Event.end_? (?= (DateAtTimeWithDefaults (Now) (NumberPM 2L))))))) 
\end{lstlisting} 
& Can you make it end at 2 pm? 
\\
	\bottomrule
\end{tabular}

    \end{adjustbox}
         \caption{Sample parse-trees and corresponding (conditioned) utterances from the 2-stage method, with and without differential privacy.}
        \label{tab:trees_gens}
\end{table*}

\subsection{Privacy Accounting}\label{app:privacy}
We chose in Algorithm~\ref{alg:2stage} (\S\ref{sec:2stage}) to split $T$ into $T_1$ and $T_2$ and share the other DP parameters across the two stages.  This lets us use a single shared moments accountant~\cite{abadi2016deep} and thus benefit from sub-linear composition. 

To be precise, if our method for ensuring $(\epsilon,\delta)$-DP happens to guarantee $(\epsilon_i,\delta)$-DP at each stage $i$ (as a result of training for $T_i$ epochs with clipping threshold $C$, batch size $|B|$, and noise multiplier $\sigma$), then we have $\epsilon \leq \epsilon_1+\epsilon_2$.  We may enjoy $\epsilon < \epsilon_1+\epsilon_2$ \cite{kairouz2015composition,mironov2017renyi}.  

Thus, we are in general able to train for more total epochs, or with lower noise multipliers, than if we had directly divided the privacy budget as $\epsilon=\epsilon_1+\epsilon_2$, enforced $(\epsilon_i,\delta)$-DP at each stage $i$ to guarantee $(\epsilon,\delta)$-DP overall by linear composition, and used that commitment to determine the maximum allowed $T_i$ (for a given $\sigma_i$) or the minimum allowed $\sigma_i$ (for a given $T_i$) at each stage $i$.

\section{Additional Related Work}\label{app:related-work}

In this section, we discuss additional related work beyond Section \ref{sec:related-work}.

\subsection{Differentially Private Training and Synthesis}

In our work, we took samples from a generative language model trained with differential privacy~\cite{abadi2016deep,li2021large,yu2021differentially,kerrigan-etal-2020-differentially,shi2021selective,anil2021large,tian2021seqpate}, to build a synthesized dataset which would then be used to improve different down-stream tasks.

As an alternative, it is also possible to decode in a differentially private way from a non-DP model~\cite{ginart2022submix}
Other work has proposed DP $n$-grams~\cite{kim2021differentially}, which helps extract common $n$-grams from the data privately. $n$-grams, however, are not much help in improving performance on downstream tasks (parsing or NLU classification), unless $n$ is quite large.

\subsection{Semantic Parsing}

Other than \citet{yang-etal-2022-addressing},
we are not aware of other works that specifically address learning from private data for semantic parsing. 
Nevertheless, many papers have explored ways to construct, improve, and adapt semantic parsing systems with minimal amounts of supervision.
\citet{wang-etal-2015-building} construct semantic parsing datasets by first enumerating potential parse trees (as ``canonical utterances'') and then asking crowd workers to convert them into utterances.
\citet{su-yan-2017-cross} leverage the canonical utterances for cross-domain generalization in semantic parsing by reformulating semantic parsing as paraphrasing from input utterances to canonical utterances.
\citet{yin-etal-2022-ingredients} is a later instantiation of a similar idea, with automated paraphrasing of canonical utterances into natural utterances and a few other components.
\citet{zhao2019data,zhong2020grounded,cao2020unsupervised,burnyshev2021single,kim2021neuralwoz,tseng2021transferable} are other works in a similar vein.
Since these tend to use similar primitives as our 2-stage approach, but trained without DP on different sources of data, our approach is largely complementary and can be used to augment the prior approaches.

\end{document}